\documentclass{article}
\usepackage{multirow}
\usepackage{times}
\usepackage{url}            
\usepackage{hyperref}
\usepackage{amsfonts}       
\usepackage{nicefrac}       
\usepackage{tikz,pgfplots}
\usepackage{amssymb, bbm, amsmath}
\usepackage{enumitem}

\usepackage{extarrows}

\usepackage{times}

\usepackage{soul}
\usepackage{url}
\usepackage[utf8]{inputenc}
\usepackage[small]{caption}
\usepackage{amsmath}

\newtheorem{theorem}{Theorem}

\newtheorem{lemma}{Lemma}

\newtheorem{definition}{Definition}

\newtheorem{remark}{Remark}

\newcommand{\E}{\mathbb{E}}
\newcommand{\prob}{\mathbb{P}_t}
\newcommand\MV{\mathrm{MV}}

\newcommand\I{\mathbb{I}}

\usepackage{microtype}
\usepackage{graphicx}
\usepackage{subfigure}
\usepackage{booktabs} 

\usepackage[accepted]{icml2020}

\icmltitlerunning{Thompson Sampling Algorithms for Mean-Variance Bandits}

\begin{document}

\twocolumn[
\icmltitle{Thompson Sampling Algorithms for Mean-Variance Bandits}




\begin{icmlauthorlist}
\icmlauthor{Qiuyu Zhu}{iora}
\icmlauthor{Vincent Y. F. Tan}{iora,ece,math}
\end{icmlauthorlist}

\icmlaffiliation{ece}{Department of Electrical and Computer Engineering, National University of Singapore, Singapore}
\icmlaffiliation{iora}{Institute of Operations Research and Analytics, National University of Singapore, Singapore}
\icmlaffiliation{math}{Department of Mathematics, National University of Singapore, Singapore}

\icmlcorrespondingauthor{Qiuyu Zhu}{qiuyu\_zhu@u.nus.edu}
\icmlcorrespondingauthor{Vincent Y. F. Tan}{vtan@nus.edu.sg}

\icmlkeywords{Multi-armed bandit}

\vskip 0.3in
]



 \printAffiliationsAndNotice{}  

\begin{abstract}
The multi-armed bandit (MAB) problem is a classical learning task that exemplifies the exploration-exploitation tradeoff.  However, standard formulations do not take into account {\em risk}. In online decision making systems, risk is a primary concern. In this regard, the mean-variance risk measure is one of the most common objective functions. Existing algorithms for mean-variance optimization in the context of MAB problems have unrealistic assumptions on the reward distributions. We develop Thompson Sampling-style algorithms for mean-variance MAB and provide comprehensive regret analyses for Gaussian and Bernoulli bandits with fewer assumptions. Our algorithms achieve the best known regret bounds for mean-variance MABs and also attain the information-theoretic bounds in some parameter regimes. Empirical simulations show that our algorithms significantly outperform existing LCB-based algorithms for all risk tolerances.
\end{abstract}

\section{Introduction}
The MAB problem studies the problem of online learning with partial feedback. This problem has a large number of real-world applications, such as online advertising,  clinical trials, and financial portfolio design. The most widely-used MAB model is the stochastic MAB model. A player chooses among several arms, each defined by an independent reward distribution. In each period, the player plays one arm and obtains a random reward observation from that arm. The player faces a dilemma between exploiting the current information by playing the arm with highest estimated reward and exploring all arms to collecting reward information. The primary concern of this body of literature is to find a learning algorithm which can maximize the expected cumulative reward. However, scant attention has been paid to \textit{risk}. In many practical problems,  such as clinical trials, an algorithm that yields a lower expected payout but is less risky may be preferable.

 To date, there has been little agreement on the definition of risk. For example, under the MAB setting, a solution with guarantees over multiple runs of an algorithm may not satisfy the desire for a solution with low variability over a single implementation of an algorithm. 
Indeed, there are various risk modeling paradigms, such as, the {\em expected utility theory} and the {\em mean-variance paradigm}. In this paper, we focus on the mean-variance paradigm, which was introduced by \citet{markowitz1952portfolio}. We seek to understand the role of risk in the mean-variance MAB problem. 
 \subsection{Related Work}
 Although problems involving bandits have a long history, dating back to \citet{thompson1933likelihood}, risk-aware bandits problem have been studied only recently. \citet{even2006risk} incorporated risk into online learning problem. They initiated the investigation of explicit risk considerations in the standard models of worst-case online learning by considering the \textit{Sharpe ratio} and the \textit{mean-variance} criterion. 
 \citet{audibert2009exploration} analyzed the expected regret and its distribution, showed that an anytime UCB and UCB-V might have large regret with non-negligible probability. \citet{salomon2011deviations} further extended this result by showing that no anytime algorithm can achieve a regret with both a small expected regret and exponential tails. 
 These result are important steps towards the analysis of risk in MAB problems, but they are limited to the case that an algorithm's objective is to find the arm with the highest expectation. 

\citet{sani2012risk} considered risk-aversion in MAB problems. In particular, they studied the mean-variance risk criteria and presented the MV-LCB algorithm. \citet{vakili2015mean} and \citet{vakili2016risk} considered mean-variance minimization under a regret minimization framework and completed the regret analysis of \citet{sani2012risk}. In the best arm identification setting, \citet{david2016pure} and \citet{david2018pac} studied VaR-based risk criteria. Moreover, some coherent risk measures---for example  CVaR---have been studied by \citet{kolla2019risk}, \citet{xu2018index}.  \citet{galichet2013exploration} presented the MaRaB algorithm which uses CVaR$_\alpha$ in its implementation. However, they analyzed the regret under the assumptions that $\alpha = 0$ and that the CVaR$_\alpha$ and average optimal arms coincide. \citet{maillard2013robust} proposed and analyzed RA-UCB which is based on the measure of entropic risk with a parameter $\lambda$. 
\citet{cassel2018general} proposed a general approach for MAB under risk criteria, they used \textit{Empirical Distribution Performance Measures} as the performance metric of the algorithm. They presented and analyzed the U-UCB algorithm. All algorithms above are based on UCB or LCB ideas. To the best of our knowledge, there is no work on using Thompson Sampling in MAB whilst incorporating risk. 

Another line of research concerns variations of the assumptions on the reward distributions. \citet{bubeck2013bandits} showed that finite moments of order 2 (i.e. finite variance) are sufficient to obtain regret bounds of the same order as if one assumes sub-Gaussian rewards. \citet{liu2011multi} proposed the DSEE approach to complement existing work on MAB by providing results under a set of relaxed conditions on the reward distributions. \citet{yu2018pure} investigated the problem on pure exploration of MAB with heavy-tailed payoffs by relaxing the assumption of payoffs with sub-Gaussian noises. These works show that it is possible to achieve meaningful regret bounds when the reward distribution is not sub-Gaussian. 
For example, square of Gaussian has a $\chi^2$ distribution, which is not sub-Gaussian; hence, the analyses of these works are not applicable.
\subsection{Contributions}
In this paper, we focus on the MABs under the mean-variance risk criterion. Our contributions are as follows:

\begin{itemize}[leftmargin=*]
\item  \textbf{Four algorithms:} We propose three Thompson Sampling-based algorithms for Gaussian bandits---MTS, VTS, and MVTS---each suitable for use in different regimes. We also demonstrate flexibility  and generality of our approach by proposing  BMVTS, another Thompson-Sampling-based algorithm, but this time, for Bernoulli bandits. 
\item  \textbf{Comprehensive regret analyses:} We provide theoretical analyses of the algorithms and show that in some regimes, the regrets either generalize existing results \cite{agrawal2012analysis} or meet the information-theoretic lower bound by \citet{vakili2016risk}. The analyses are novel because previous methods for risk-averse MAB problems impose a sub-Gaussian assumption on the variance of the reward distributions so are not applicable to our Gaussian setting. Thus, we need to derive new anti-concentration bounds (cf.\ Lemmas~\ref{lemma: lem3} and~\ref{lemma: lem4}). The regret is also analyzed for BMVTS.
\item \textbf{Extensive set of simulations:} We provide extensive sets of simulations for both Gaussian and Bernoulli bandits to show that our algorithms outperform state-of-the-art algorithms for mean-variance bandits. In particular,  MVTS dominates  LCB-based algorithms over all risk tolerances~$\rho$.
\end{itemize}

In the majority of the paper, we use Gaussian bandits as an example to illustrate our algorithms and proof techniques, but the same method {(albeit with different concentration bounds)} can be use to prove regret bounds for Bernoulli bandits (cf.\ Theorem~\ref{thm: thm4}).

The rest of this paper is organized as follows. 
 We introduce mean-variance MABs and some notations in Section~\ref{sec: formulation}. In Section~\ref{sec: algorithms}, we present Thompson Sampling algorithms for mean-variance Gaussian bandits. Some regret analyses are provided in Section~\ref{sec: analysis}. A set of numerical simulations is reported to validate the theoretical results in Section~\ref{sec: simulation}. In Section~\ref{sec:conc}, we conclude the discussions. 
Detailed/full proofs are deferred to the supplementary material. 
\section{Problem formulation}
\label{sec: formulation}
In this section we introduce the main notations and define mean-variance MABs. Consider a MAB $\nu$ with $K$ arms and a single player.  The problem is defined over a time horizon of length $n$. The mean and the variances of the reward distributions are fixed and unknown (the frequentist setting). At each time $t \in\{1,\ldots,n\}$, the player chooses one arm to play. Playing arm $i$ at time $t$ yields a random reward $X_{i, t}$ drawn from $\nu_i$. All reward samples are independent conditionally to the choice of the arm. A {\em policy} $\pi\left(\cdot\right):(t, A_1, X_1, \ldots, A_{t-1}, X_{t-1})\rightarrow [K]$, is a function that specifies the action of the player at each time. The policy depends on the history $(A_1, X_1, \ldots, A_{t-1},X_{t-1})$. Let $T_{i, n}$ denote the number of times that the player pulls arm $i$ during the time periods $\{1,\ldots,n\}$. 

In the standard MAB problem, the objective of the player is to minimize the expected cumulative regret. 
Here, instead, we focus on finding the arm which effectively balances its expected reward and variability. Although there are a large number of models for the trade-off between return and risk, such as {\em Sharpe ratio} from \citet{sharpe1966mutual} and the {\em Knightian uncertainty} from \citet{knight1921risk}, here we focus on the most popular and simple model, namely, the mean-variance model proposed by \citet{markowitz1952portfolio}. 
\begin{definition}\label{def: def1}
	The {\em mean-variance} of an arm $i$ with mean $\mu_i$, variance $\sigma_i^2$ and risk tolerance $\rho$ is  $\MV_i =\rho\mu_i - \sigma_i^2$.
\end{definition}

Let arm $1$ be the best arm, i.e., $\MV_1 = \max_{i\in [K]} \MV_i$. Based on Definition~\ref{def: def1}, we can recover two extreme cases by considering the extremal values of the  risk tolerance $\rho$. When $\rho=0$, the mean-variance MAB is a variance minimization problem. As $\rho\rightarrow\infty$, the problem reduces to standard MAB in which one seeks to maximize the reward. 

Given i.i.d.\ samples $\{X_{i, s}\}_{s=1}^t$ drawn from distribution $\nu_i$, we define the empirical mean-variance as
\begin{align}
	\widehat{\MV}_{i, t}& = \rho \hat{\mu}_{i, t} - \hat{\sigma}_i^2,\qquad\mbox{where}  \label{equ: equ1}\\
\hat{\mu}_{i, t} &= \frac{1}{t}\sum_{s=1}^t X_{i, s}, \quad\hat{\sigma}_{i,t}^2=\frac{1}{t}\sum_{s=1}^t\left(X_{i, s}-\hat{\mu}_{i, t}\right)^2 \label{eqn:emp_var} .
\end{align}

Let $\Gamma_{i, j} = \mu_{i} - \mu_j$ and $\Delta_{i} =  \MV_1 - \MV_i$ denote, respectively, the difference between the means of arms $i$ and   $j$ and the difference between the mean-variances of   $i$ and   $1$. Let $T_{i, j}$ be the number of times that arm $i$ is pulled during first $j$ periods. Assume throughout that   $\Delta_i >0 $. We remark that using the unbiased estimate of the sample variance $\hat{\sigma}_i^2=\frac{1}{t-1}\sum_{s=1}^t\left(X_{i, s}-\hat{\mu}_{i, t}\right)^2$ does not change our regret bound and conclusions. Hence, we use the definition of the sample variance in~\eqref{eqn:emp_var} for simplicity.

Given a learning policy $\pi (\cdot)$ and the reward process $\{X_{\pi(t), t}\}_{t=1}^n$, we define the {\em empirical mean-variance} as 
    \begin{align}
    \label{equ: equ3}
    \widehat{\MV}_n\left(\pi\right) & = \rho \hat{\mu}_n\left(\pi\right) - \hat{\sigma}_n^2(\pi) ,\quad\mbox{where}\\
    	\hat{\mu}_n (\pi) &= \frac{1}{n}\sum_{t=1}^n X_{\pi(t), t}, \quad\mbox{and}\\
    	 \hat{\sigma}_n^2(\pi) &= \frac{1}{n}\sum_{t=1}^n\left(X_{\pi(t), t}-\hat{\mu}_n\left(\pi\right)\right)^2.
    \end{align}
Obviously, the optimal policy should choose arm $1$ for all $t\in\{1,\ldots, n\}$. For each policy $\pi$, this leads to the definition of the regret, which is the difference of the empirical mean-variance of the policy and the optimal mean-variance. 
\begin{definition} \label{def: def2}
	The {\em expected regret} of a policy $\pi(\cdot)$ over $n$ rounds is defined as 
	\begin{equation}
	\E\left[\mathcal{R}_n\left(\pi\right)\right] = n\big(\MV_1-\E \big[\widehat{\MV}_n(\pi )\big]\big). \label{eqn:exp_reg}
	\end{equation}
\end{definition} 
 We remark the expectation in \eqref{eqn:exp_reg} is taken over the sample path
of the rewards $\{X_{\pi(t),t}\}_{t=1}^n$.  The expected regret can alternatively be written as the expectation of
\begin{align}
	\mathcal{R}_n\left(\pi\right)&=  \mathcal{R}_n^{(1)}\left(\pi\right) + \mathcal{R}_n^{(2)}\left(\pi\right), \quad\mbox{where}   \nonumber\\
	\mathcal{R}_n^{(1)}\left(\pi\right) &\!:= \!n\sum_{i=1}^K \left(\!\rho\frac{T_{i,n}}{n}(\mu_1\! -\! \hat{\mu}_{i,T_{i,n}})\!  +\!  \frac{T_{i,n}}{n}(\hat{\sigma}_{i,T_{i,n}}^2\! -\! \sigma_1^2)\!\right),\nonumber\\
\mathcal{R}_n^{(2)}\left(\pi\right)& \!:=\! n\sum_{i=1}^K\frac{T_{i,n}}{n}(\hat{\mu}_{i,T_{i,n}}-\hat{\mu}_n(\pi))^2 . \nonumber
\end{align}
  This definition of expected regret leads to a natural objective---to design an algorithm whose regret increases as slowly as possible as $n$  increases. The objective of our problem is to balance the tradeoff between risk and return, but this definition of regret does not give us a view of how the components of the regret (i.e.,  the regret related to the risk and regret related to the return) influences the overall regret $\E\left[\mathcal{R}_n\left(\pi\right)\right] $. This motivates the following quantity. 
\begin{definition} \label{def: def3}
	The {\em  expected pseudo-regret} for a policy $\pi(\cdot)$ over $n$ rounds is defined as 
\begin{equation}
\label{equ: pseudoregret}
\!\E\big[\widetilde{\mathcal{R}}_n(\pi)\big] \!=\! \sum_{i=2}^K\! \E\left[T_{i, n}\right]\Delta_i+\frac{1}{n}\!\sum_{i=1}^K\sum_{j\neq i} \! \E\left[T_{i, n}T_{j, n}\right]\Gamma_{i, j}^2.\!\end{equation}
\end{definition}
 Let us relate the expected regret with the expected pseudo-regret. 
The expected pseudo-regret can be divided into two parts. The first term in~\eqref{equ: pseudoregret}  is the regret of the expected mean-variance that results from choosing suboptimal arms, which is exactly the expectation of $\mathcal{R}_n^{(1)}\left(\pi\right)$. Given a policy $\pi(\cdot)$, the variance of the reward process $\{ X_{\pi(t),t}\}_{t=1}^n$ also influences $\E[\mathcal{R}_n^{(1)}\left(\pi\right)]$, but as can be seen from the definition of $\mathcal{R}_n^{(1)}\left(\pi\right)$, the regret of the variance $n(\hat{\sigma}_n^2(\pi)-\sigma_1^2)$ is only one of the two terms that comprises $\mathcal{R}_n^{(1)}\left(\pi\right)$. We also have to take into account the regret of variance which arises due to the switching between different arms; this  is the term $\mathcal{R}_n^{(2)}\left(\pi\right)$. The second term in~\eqref{equ: pseudoregret} is an upper bound of $\E[\mathcal{R}_n^{(2)}\left(\pi\right)]$ (see the proof of Lemma~\ref{lemma: lem1}). 
Because the number pulls of suboptimal arms $T_{i,n}$  for $i\ne 1$ is explicitly delineated in the expected pseudo-regret, this version of the regret is easier to work with in the sequel.

\begin{lemma} \label{lemma: lem1} 
The difference between the expectations of these two expected regrets can be bounded as follows:
	\begin{equation}\E\left[\mathcal{R}_n(\pi)\right] \leq \E\big[\widetilde{\mathcal{R}}_n(\pi)\big] + 3\sum_{i=1}^K \sigma_i^2.
	\end{equation}
\end{lemma}
This lemma shows that we can obtain a bound on the expected regret by proving a bound on the expected pseudo-regret. Hence, we focus on the analysis of $\widetilde{\mathcal{R}}_n(\pi)$. The second term  in \eqref{equ: pseudoregret} can be upper bounded  as follows: 
\begin{align}
	&\frac{1}{n}\sum_{i=1}^K\sum_{j\neq i} \E\left[T_{i, n}T_{j, n}\right]\Gamma_{i, j}^2 \nonumber\\
	&= \frac{1}{n}\Bigg(\sum_{j\neq 1} \E\left[T_{1,n}T_{j, n}\right]\Gamma_{1, j}^2 + \sum_{i=2}^K\sum_{j\neq i} \E\left[T_{i, n}T_{j, n}\right]\Gamma_{i, j}^2 \Bigg) \nonumber\\
	&\leq  \frac{1}{n}\Bigg(\sum_{j\neq 1} n\E\left[T_{j,n}\right]\Gamma_{1, j}^2 + \sum_{i=2}^K n\E\left[T_{i,n}\right]\Gamma_{i, \max}^2 \Bigg) \nonumber\\
	&\le 2\sum_{i=2}^K \E\left[T_{i, n}\right]\Gamma_{i, \max}^2, \label{eqn:bd_above}
\end{align}
where $\Gamma_{i, \max}^2 = \max \{ (\mu_i-\mu_j)^2: j=1,\ldots,K\}$.

By applying \eqref{eqn:bd_above}, the pseudo-regret can be written as 
\begin{equation}\E\big[\widetilde{\mathcal{R}}_n(\pi)\big]\leq \sum_{i=2}^K\E\left[T_{i, n}\right]\left(\Delta_i + 2\Gamma_{i, \max}^2\right).\label{equ: equ10}\end{equation} 
This inequality shows that it suffices to bound the  number of pulls of each suboptimal arm $i\ne 1$. 


\section{Algorithms for mean-variance MAB}
\label{sec: algorithms}
In this section, we introduce three Thompson Sampling-based risk-averse MAB algorithms. To illustrate the design of the algorithms, we consider the Gaussian bandits in detail, i.e., $\nu\in \mathcal{E}_\mathcal{N}^K(1) = \{\nu = (\nu_1,\ldots,\nu_K): \nu_i\sim\mathcal{N}(\mu_i,\sigma_i^2),\sigma_i^2\leq 1\}$. We provide short discussions concerning Bernoulli bandits in Sections~\ref{sec:Berbandits} and~\ref{sec:bernoulli}.

An important step of Thompson Sampling algorithms is the updating of parameters based on Bayes rule. Consider the general prior for the Gaussian with unknown means and precisions, i.e.,\ the Normal-Gamma prior.  Let $\mu$ and $\tau$ be the mean and precision of the Gaussian respectively. If $(\mu,\tau)\sim \mathrm{Normal\text{-}Gamma}(\mu,T,\alpha,\beta)$, then $\tau\sim \mathrm{Gamma}(\alpha,\beta)$, and $\mu|\tau \sim \mathcal{N}(\mu, 1/(\tau T))$. Since the Normal-Gamma distribution is the conjugate prior for the Gaussian with unknown mean and variance, we use Algorithm~\ref{alg: UPDATE} to update these parameters.

\begin{algorithm}[tb]
\caption{Update $(\hat{\mu}_{i, t-1}, T_{i, t-1}, \alpha_{i, t-1} , \beta_{i, t-1})$}
\label{alg: UPDATE}
\begin{algorithmic}[1]
	\STATE \textbf{Input}: Prior parameters\ $(\hat{\mu}_{i, t-1}, T_{i, t-1}, \alpha_{i, t-1} , \beta_{i, t-1})$ and  new sample $X_{i, t}$
	\STATE Update the mean: $\hat{\mu}_{i, t} = \frac{T_{i, t-1}}{T_{i, t-1}+1}\hat{\mu}_{i, t-1} + \frac{1}{T_{i, t-1}+1}X_{i, t}$
	\STATE Update the number of samples and the shape parameter: $T_{i, t} = T_{i, t-1} + 1,\alpha_{i, t} = \alpha_{i, t-1}+\frac{1}{2}$
	\STATE Update the rate parameter:
	$\beta_{i, t} = \beta_{i, t-1} + \frac{T_{i, t-1}}{T_{i, t-1}+1}\cdot \frac{(X_{i, t}-\hat{\mu}_{i, t-1})^2}{2}$
\end{algorithmic}	
\end{algorithm}


\begin{algorithm}[t]
\caption{Thompson Sampling for Mean Learning   ({MTS}) and Variance Learning (VTS)}
\label{alg: M&VTS}
\begin{algorithmic}[1]
    \STATE \textbf{Input}: $\hat{\mu}_{i, 0}=0, T_{i, 0}=0, \alpha_{i, 0} = \frac{1}{2} , \beta_{i, 0} = \frac{1}{2}$. 
	\FOR {each $t = 1, 2, \ldots, K$}{}
	    \STATE Play arm $t$ and observe the reward $X_{t, t}$ 
	    \STATE $\mathrm{Update}(\hat{\mu}_{t, t-1}, T_{t, t-1}, \alpha_{t, t-1}, \beta_{t, t-1})$
	\ENDFOR
	\FOR {each $t = K+1, \ldots$, } {}
	\STATE \textbf{MTS}: 
		\begin{itemize}
			\item Sample $\theta_{i,t}$ from $\mathcal{N}\left(\hat{\mu}_{i, t-1}, 1/T_{i, t-1}\right)$. 
		    \item Play arm $i\left(t\right)=\arg\max_{i}\rho \theta_{i,t}- 2\beta_{i, t-1}$ and observe reward $X_{i\left(t\right), t}$. 
		\end{itemize}
	\STATE \textbf{VTS}:
	\begin{itemize}
		\item  Sample $\tau_{i,t}$ from $\mathrm{Gamma}\left(\alpha_{i, t-1}, \beta_{i, t-1}\right)$. 
	    \item Play arm $i(t)=\arg\max_{i\in[K]}\rho\hat{\mu}_{i, t-1}-1/\tau_{i,t}$ and observe reward $X_{i\left(t\right), t}$. 
	\end{itemize}
		\STATE Update$(\hat{\mu}_{i(t), t-1}, T_{i(t), t-1}, \alpha_{i(t), t-1}, \beta_{i(t), t-1})$

	\ENDFOR
\end{algorithmic}	
\end{algorithm}

\subsection{Thompson Sampling algorithms for mean learning and variance learning}

We propose a variant of Thompson Sampling algorithm to solve the mean-variance Gaussian MAB problem when $\rho$ is large; we call this Mean Thompson Sampling (MTS). In each period, the variance of each arm will be estimated and the algorithm sequentially updates the posterior mean of each arm. We propose another algorithm to handle the case in which $\rho$ is small. We call this algorithm Variance Thompson Sampling (VTS).   VTS estimates the mean  of each arm and updates the posterior precision of each arm. At the beginning of each period, VTS samples a precision from the posterior and then selects the optimal arm according to the estimates of the mean and the Thompson samples of the precision. Pseudocodes of the MTS and VTS algorithms are shown in Algorithm~\ref{alg: M&VTS}.

\begin{algorithm}[t]
\caption{Thompson Sampling for Gaussian mean-variance bandits ({MVTS})}
\label{alg: MVTS}
\begin{algorithmic}[1]
	\STATE \textbf{Input}: $\hat{\mu}_{i, 0}=0, T_{i, 0}=0, \alpha_{i, 0} = \frac{1}{2} , \beta_{i, 0} = \frac{1}{2}$. 
	\FOR {each $t = 1, 2, \ldots, K$}{}
	    \STATE Play arm $t$ and update $\hat{\mu}_{t, t} = X_{t, t}$. 
	    \STATE $\mathrm{Update}(\hat{\mu}_{t, t-1}, T_{t, t-1}, \alpha_{t, t-1}, \beta_{t, t-1})$
	\ENDFOR
	\FOR {each $t = K+1, K+2, \ldots$, } {}
		\STATE Sample $\tau_{i,t}$ from $\mathrm{Gamma}(\alpha_{i, t-1}, \beta_{i, t-1})$. 
	    \STATE Sample $\theta_{i,t}$ from $\mathcal{N}(\hat{\mu}_{i, t-1}, 1/T_{i, t-1})$
	    \STATE Play arm $i(t)=\arg\max_{i\in[K]}\rho \theta_{i,t}-1/\tau_{i,t}$ and observe reward $X_{i(t), t}$
        \STATE $\mathrm{Update}(\hat{\mu}_{i(t), t-1}, T_{i(t), t-1}, \alpha_{i(t), t-1}, \beta_{i(t), t-1})$
	\ENDFOR
\end{algorithmic}	
\end{algorithm}

\subsection{A Thompson Sampling algorithm for the Gaussian mean-variance MAB}
The proposed algorithms can effectively solve the risk-averse MAB problem in two extreme scenarios (e.g.,  large or small $\rho$). However, it is difficult to decide on a suitable threshold for a player  to choose the algorithm she needs because she does not know the true means and variances (and neither does she know $\rho$). In this section, we propose a {\em combined} or {\em unified} Thompson Sampling  algorithm to address this problem. The player chooses a prior over the set of feasible bandits parameters for both the mean and precision. In each round, the player samples a pair of parameters from each posterior and plays an arm according to the optimal action under these parameters. 

The Mean-Variance Thompson Sampling  (MVTS) algorithm, shown in Algorithm~\ref{alg: MVTS}, uses the conjugate prior of a Gaussian which is parametrized by the mean and precision. When $\rho$ is small, the mean-variance MAB problem reduces to a variance minimization problem. In this regime,  {MVTS} is consistent with {VTS}. On the other hand, when $\rho\rightarrow\infty$, the problem reduces to the standard reward maximization problem. In this case,  {MVTS} is   consistent with {MTS}. 
Hence, we expect that  {MVTS} performs well over all $\rho\in\mathbb{R}_+$. We show that this is indeed the case in Theorem \ref{thm: thm3}. 

\begin{algorithm}[t]
\caption{Thompson Sampling for Bernoulli mean-variance bandits ({BMVTS})}
\label{alg: BMVTS}
\begin{algorithmic}[1]
	\STATE \textbf{Input}: $\alpha_{i,1}=1,\beta_{i,1}=1$. 
	\FOR {each $t = 1, 2, \ldots$, } {}
		\STATE Sample $\theta_{i,t}$ from $\mathrm{Beta}(\alpha_{i,t},\beta_{i,t})$. 
	    \STATE Play arm $i(t)=\arg\max_{i\in[K]}\rho \theta_{i,t}- \theta_{i,t}(1-\theta_{i,t})$ and observe reward $X_{i(t), t}$
        \STATE Update parameters:\ $\alpha_{i,t+1} = \alpha_{i,t} + X_{i(t),t},$
         \\$\beta_{i,t+1} = \beta_{i,t} + (1-X_{i(t),t})$
	\ENDFOR
\end{algorithmic}	
\end{algorithm}
\subsection{A Thompson sampling algorithm for Bernoulli mean-variance bandits}\label{sec:Berbandits}
In this section, we present the BMVTS algorithm for Bernoulli mean-variance bandit problem, which is BMVTS. Under Bernoulli bandits setting, the reward of arm $i$ is generated from Bernoulli distribution with success probability $p_i$. Hence, we use the Beta distribution as the prior of each arm's reward distribution. Pseudocode of the BMVTS algorithm is shown in Algorithm \ref{alg: BMVTS}.

\section{Regret analysis}
\label{sec: analysis}
We present our regret bounds and a sketch of the proof for MVTS. 
There are some non-trivial technical details that are required for MVTS because of the random variables involved in some error events; see the dependencies of the random variables in Figure~\ref{fig: structure}, Lemmas \ref{lemma: lem3} and~\ref{lemma: lem4} and their accompanying discussions. The {\em asymptotic} regret bounds that we present here are derived from the {\em finite-horizon} regret bounds, which are available in the supplementary materials  (see Theorem \ref{thm: S-1}, \ref{thm: S-2}, \ref{thm: S-3}).

\subsection{Regret analysis for MTS}
	\begin{theorem} \label{thm: thm1}
		If $\rho > \max\big\{\sigma_1^2 / \Gamma_i : i=1, 2, \ldots, K\big\}$, the asymptotic expected regret produced by MTS   for mean-variance Gaussian bandits satisfies
		\begin{equation}
		\varlimsup_{n\rightarrow \infty}\!\frac{\E\big[\mathcal{\widetilde{R}}_n\left(\mathrm{MTS}\right)\big]}{\log n} \le\sum_{i=2}^K \frac{2\rho^2}{ (\rho\Gamma_{1,i}-\sigma_1^2)^2}\left(\Delta_{i}\!+\! 2\Gamma_{i, \max}^2\right).\end{equation}
	\end{theorem}
By  Lemma \ref{lemma: lem1}, the same   bound holds for $\E[\mathcal{R}_n(\mathrm{MTS})]$.  This remark also applies to Theorems~\ref{thm: thm2} and~\ref{thm: thm3}.
\begin{remark}[The assumption]
	\em The reason for the assumption on $\rho$ is that the mean-variance has the same order as the mean of each arm. Thus the mean is the dominant term. However, in our numerical simulations, we can observe that MTS still performs well even this condition is not met in practice (see Figure \ref{fig: fig3}).
\end{remark}

\subsection{Regret analysis for VTS}
\begin{theorem} \label{thm: thm2}
		Let $h(x) = \frac{1}{2}(x-1-\log x) $. If $\rho\leq \min\big\{{\Delta_i}/{\Gamma_i}: {\Delta_i}/{\Gamma_i} >0\big\}$  and  $\Gamma_i^2> 2\sigma_1^2h\left({\sigma_i^2}/{\sigma_1^2}\right)$ for all $i,$  the asymptotic expected regret of VTS for mean-variance Gaussian bandits satisfies 
		\begin{equation}\varlimsup_{n\rightarrow \infty}\frac{\E \big[\mathcal{\widetilde{\mathcal{R}}}_n\left(\mathrm{VTS}\right)\big]}{\log n} \leq \sum_{i=2}^K \frac{1}{ h\left( {\sigma_i^2}/{\sigma_1^2}\right)}  \left(\Delta_i+2\Gamma_{i, \max}^2\right). \label{eqn:bd_vts}\end{equation}
\end{theorem}

\begin{remark}[Assumptions] {\em \citet{bubeck2013bandits} show that the only condition we need to achieve the same form of regret bound as the sub-Gaussian case is that the reward distributions have finite variance. Here, we consider   Gaussian bandits; hence this assumption is fulfilled, leading to the bound $\E  [\mathcal{ {\mathcal{R}}}_n (\mathrm{VTS} ) ] =O(\log n)$. However, other works on risk-averse MABs require more stringent conditions on $\nu$, e.g., \citet{vakili2015mean} assume that the empirical variance of $\nu_i$ is sub-Gaussian and~\citet{sani2012risk} use the assumption that the reward is bounded almost surely.}
\end{remark}

\begin{remark}[Scale invariance]
{\em The bound  in~\eqref{eqn:bd_vts}  depends only on the ratio of the variances, which is similar to the fact that the regret for the standard MAB depends only on the differences of the means. 
This justifies our assumption that $\nu\in \mathcal{E}_\mathcal{N}^K(1)$ since we can rescale the variances.}
\end{remark}

\subsection{Regret analysis for MVTS}
\begin{theorem} \label{thm: thm3}
		 The asymptotic expected regret of MVTS for mean-variance Gaussian bandits satisfies 
		\begin{multline}\varlimsup_{n\rightarrow \infty}\frac{\E \big[\mathcal{\widetilde{\mathcal{R}}}_n\left(\mathrm{MVTS}\right)\big]}{\log n} \\
		\;\;\leq \sum_{i=2}^K \max\bigg\{\frac{2}{\Gamma_{1,i}^2},\frac{1}{h( {\sigma_i^2}/{\sigma_1^2})}\bigg\} \left(\Delta_i+2\Gamma_{i, \max}^2\right). \label{eqn:bd_vts1}\end{multline}
\end{theorem}

\begin{remark}[The extreme cases] \label{rmk:extreme}
{\em When take $\rho$ to its extremal values, we can elaborate on discussion after Definition~\ref{def: def1} in terms of the regrets. 
First, as $\rho\rightarrow \infty$, the mean will dominate the mean-variance. Hence, $\prob(\widehat{\MV}_{i,t}\geq \MV_1-(\rho+1)\varepsilon) \rightarrow\prob(\theta_{i,t}\geq \mu_1-\varepsilon)$. By applying the same proof technique as that for MVTS, so we have 
\[\varlimsup_{n\rightarrow \infty}\varlimsup_{\rho\rightarrow \infty}\frac{\E \big[\mathcal{\widetilde{\mathcal{R}}}_n\left(\mathrm{MVTS}\right)\big]}{\rho\log n} 
		\leq \sum_{i=2}^K \frac{2}{\Gamma_{1,i}}. \label{eqn:bd_mts}\]
This shows us that MVTS has a similar behavior as MTS when $\rho$ is large. When $\rho\rightarrow 0$, we have $\Delta_i \rightarrow \sigma_i^2 - \sigma_1^2$,  $ \prob(\widehat{\MV}_{i,t}\geq \MV_1-(1+\rho)\varepsilon) \rightarrow \prob(-\frac{1}{\tau_{i,t}}\geq -\sigma_1^2-\varepsilon)$, so that
\[\varlimsup_{n\rightarrow \infty}\varlimsup_{\rho\rightarrow 0}\frac{\E \big[\mathcal{\widetilde{\mathcal{R}}}_n\left(\mathrm{MVTS}\right)\big]}{\log n} \leq \sum_{i=2}^K \frac{\sigma_i^2-\sigma_1^2+2\Gamma_{i, \max}^2}{h( {\sigma_i^2}/{\sigma_1^2})}  .\] 
Thus, MVTS can learn the variances. Compare this to \eqref{eqn:bd_vts} and note that $\Delta_i\to\sigma_i^2-\sigma_1^2$ so when $\rho\to 0$, the problem reduces to the variance minimization problem. 
		These conclusions are corroborated by our numerical simulations. }
	\end{remark}
\subsection{Regret analysis for BMVTS} \label{sec:bernoulli}
\begin{theorem} \label{thm: thm4}
If $\rho \in (0,1)$, then the  asymptotic expected regret of BMVTS for mean-variance Bernoulli bandits satisfies 
\begin{align}
&\varlimsup_{n\rightarrow \infty}\frac{\E \big[\mathcal{\widetilde{\mathcal{R}}}_n\left(\mathrm{BMVTS}\right)\big]}{\log n}  \label{eqn:ber}\\ 
		&\;\;\leq \sum_{i=2}^K \max\left\{\frac{1}{2\Gamma_i^2},\frac{1}{2\left(1\! -\! \rho\! -\! p_1\! -\! p_i\right)^2}\right\} \left(\Delta_i\! +\! 2\Gamma_{i, \max}^2\right). \nonumber
\end{align}
\end{theorem}

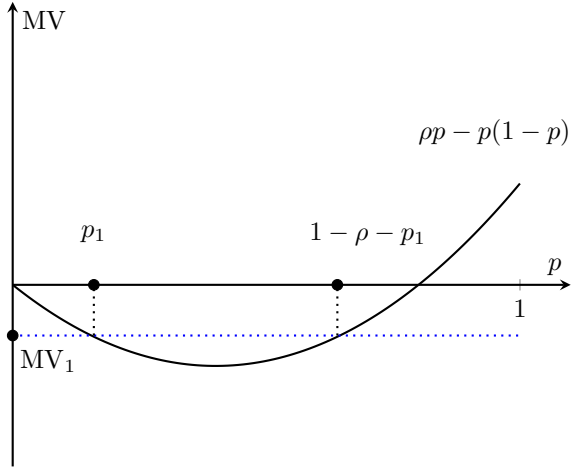
\begin{figure}[t]
\begin{center}
\begin{tikzpicture}[baseline]
\begin{axis}[
axis y line=center,
axis x line=middle,
axis line style = thick,
axis equal,
xmax=1.1,xmin=0,
ymin=-0.2,ymax=0.4,
xlabel=$p$,ylabel=$\mathrm{MV}$,
xtick={0,1},
ytick={0},
width=9cm,
anchor=center,
samples = 100,
]
\addplot[thick,domain=0:1]{x^2-0.8*x} ;
\addplot +[mark=none,blue,dotted,thick
] coordinates {(0, -0.1) (1, -0.1)};
\addplot +[mark=none,black,dotted,thick
] coordinates {(0.16, -0.1) (0.16, 0)};
\addplot +[mark=none,black,dotted,thick
] coordinates {(0.64, -0.1) (0.64, 0)};
\node[] at (axis cs: 0.95,0.3) {$\rho p-p(1-p)$};
\node[] at (axis cs: 0.07,-0.15) {$\MV_1$};
\node[] at (axis cs: 0.16,0.1) {$p_1$};
\node[] at (axis cs: 0.7,0.1) {$1-\rho-p_1$};
\addplot [only marks] table {0.16 0
0.64 0
0 -0.1
};
\end{axis}
\end{tikzpicture}
\caption{Mean-variance of Bernoulli bandit}
\label{fig:ber}
\end{center}
\end{figure}

\begin{figure}[t]
	\begin{picture}(300,90)(10,80)
		\thicklines
		\put(75,55){\framebox(110,20){$\widehat{\MV}_{i,t} = \rho\theta_{i,t}-1/\tau_{i,t}$}}
		\put(140,95){\framebox(110,20){$\tau_{i, t}\sim \mathrm{Gamma}(\alpha_{i,t}, \beta_{i,t})$}}
		\put(10,95){\framebox(110,20){$\theta_{i,t} \sim\mathcal{N}\left(\hat{\mu}_{i, T_{i,t}}, 1/ T_{i,t}\right)$}}
		\put(10,135){\framebox(110,20){$\widehat{\mu}_{i,T_{i,t}}\sim \mathcal{N}(\mu_i,\sigma_i^2/T_{i,t})$}}
		\put(140,135){\framebox(110,20){$2\beta_{i,t}/\sigma_i^2\sim \chi^2_{s-1}$}}
		\put(60,135){\vector(0,-1){20}}
		\put(190,135){\vector(0,-1){20}}
		\put(60,95){\vector(2,-1){40}}
		\put(210,95){\vector(-2,-1){40}}
	\end{picture}
	\vspace{.3in}
	\caption{Hierarchical structure of the   mean-variance Thompson samples in MVTS.}
	\label{fig: structure}
\end{figure}
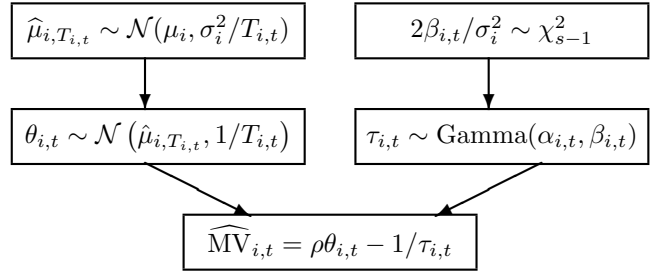

\begin{remark}[The asssumption on $\rho$]
{\em If $\rho\geq 1$, the mean-variance   $\MV(p) = \rho p- p(1-p)$ is increasing in $p\in [0,1]$, reducing  to a standard MAB. Hence, we   consider $\rho\in (0,1)$.  }
\end{remark}

\begin{remark}[The bound]
{\em The mean-variance of the best arm is $\MV_1 = \rho p_1- p_1(1-p_1)$. Hence, this value of the mean-variance  $\MV_1 $ corresponds to {\em two} different $p_1$'s (i.e., $p_1$ and $1-\rho-p_1$). See Fig.~\ref{fig:ber}.  This introduces two error events in which one suboptimal arm performs better than the best arm; this is the reason for the appearance of the ``max'' in~\eqref{eqn:ber}.}
\end{remark}

\subsection{Comparisons with lower bounds}
\label{sec: comparison}
\subsubsection*{Regret bound in Theorem \ref{thm: thm1}}
Recall that $\Delta_i=\sigma_i^2-\sigma_1^2+\rho\Gamma_{1,i} $, so that in the regime in which $\rho\to+\infty$, $\Delta_i=\Theta(\rho\Gamma_{1,i})$ and  we observe that 
\begin{equation}\lim_{\rho\rightarrow \infty}\varlimsup_{n\rightarrow \infty}\frac{\E\big[\widetilde{\mathcal{R}}_n\left(\mathrm{MTS}\right)\big]}{\rho\log n} \leq \sum_{i=2}^K \frac{2}{\Gamma_{1,i}}\label{equ: equ9} . \end{equation}
This is consistent with the fact that for $\rho\rightarrow \infty$, the mean-variance problem reduces to the maximization of the reward (without the risk aspect), for which Thompson Sampling is already known to be nearly-optimal (\citet{agrawal2012analysis}). The bound in $(\ref{equ: equ9})$ coincides with Theorem 36.3 in \citet{lattimore2018bandit}. Hence Theorem \ref{thm: thm1} generalizes the analysis to all  $\rho < \infty$.
\subsubsection*{Order optimality of Theorem \ref{thm: thm2}}
\citet{vakili2015mean} proved that the expected regret of any consistent algorithm for mean-variance MAB is $\Omega\big(\frac{\log n}{\Delta^2}\big)$ where $\Delta=\min_{i\ne 1}\Delta_i$. In Theorem \ref{thm: thm2}, by Taylor's theorem, $h(x) =  (x-1)^2/4 + o( (x-1)^2) $ as $x\to1$. Consider the regime in which for all $i\ne 1$,   $ \sigma_i^2\rightarrow\sigma_1^2$ and  $\rho= o(\sigma_i^2-\sigma_1^2)$ as $\sigma_i^2-\sigma_1^2\to 0$. In this case, $\Delta_i =\Theta(\sigma_i^2-\sigma_1^2)$ and  one has $h(\sigma_i^2/\sigma_1^2)=\Theta(\Delta_i^2)$ as $\Delta_i\to 0$. Further,  since $\Gamma_{i,\max}=\Theta(1)$ for all $i$, the upper bound in  \eqref{eqn:bd_vts} reduces to $O( \frac{1}{\Delta^2})$ and so the expected regret scales as $O\big(\frac{\log n}{\Delta^2}\big)$, asymptotically matching the lower bound in~\citet{vakili2015mean}. Thus, in this particular regime, VTS is {\em information-theoretically optimal}. This regime can be thought of as a ``hard instance'' for variance learning/minimization since all the $\sigma_i$'s are close to one another and $\rho$ is asymptotically smaller than $\sigma_i^2-\sigma_1^2$; the latter implying that the importance of the mean part of the mean-variance objective is  de-emphasized.

By Remark \ref{rmk:extreme}, MVTS particularizes to MTS and VTS when $\rho\to \infty$ and $\rho \to   0^+$ respectively. According to the above discussions, we   conclude that MVTS is   order optimal when {$\rho$ assumes these extremal values.}

\subsection{Proof Sketch of Theorem~\ref{thm: thm3}}
We use MVTS as an example to demonstrate the key steps of the proofs. There are two  main difficulties in proving an upper bound of the regret for MVTS. First, the Thompson sample of precision is not sub-Gaussian. Hence, we need to derive  sufficiently tight lower and upper bounds on the tail probability of the posterior distribution (which is not sub-Gaussian). The other difficulty is the hierarchical structure of the parameters in MVTS, which is more complicated than standard MAB (see Figure~\ref{fig: structure}). We note that the Thompson samples of the mean and precision are from different distributions. Deriving upper and lower tail bounds for $\widehat{\MV}_{i,t}$ is a major challenge because the normal distribution is sub-Gaussian, but the Gamma is only sub-exponential.
 
From~\eqref{equ: equ10}, we know that it suffices to prove an upper bound of $\E\left[T_{i, n}\right]$. Let the Thompson sample of the mean-precision pair of arm $i$ for time $t$ be $(\theta_{i,t}, \tau_{i,t})$. Let the sample mean-variance be $\widehat{\MV}_{i,t} = \rho \theta_{i,t} - 1/\tau_{i,t}$. Define 
\begin{align*}
E_i(t) := \big\{\widehat{\MV}_i(t)\leq \MV_1 -(1+\rho)\varepsilon \big\}
\end{align*}
  which is the event that the Thompson sample of arm $i$ is $(1+\rho)\varepsilon$-smaller than the optimal arm at period $t$. Event $E_i(t)$ is highly likely to occur when the algorithm has explored sufficiently since arm $1$ is optimal. However, the algorithm does not choose arm $i$ when $E_i(t)$ occurs with high probability because it chooses the arm with the maximal mean-variance in each period. The algorithm chooses arm $i$ when $E_i^c(t)$, an event with small probability (under Thompson sampling), occurs. The expectation can be divided into two parts as follows. 
\begin{lemma}[\citet{lattimore2018bandit}] \label{lemma: lem2} Let $\prob(\cdot) = \mathbb{P}(\cdot|A_1,X_1,\ldots,A_{t-1},X_{t-1})$ be the probability measure conditioned on the history up to time $t-1$ and $G_{is} = \prob\left(E_i(t)^c|T_{i,t}=s\right)$. Then,
	\begin{equation}
	\label{equ: equ15}
\!\!	\E[T_{i, n}]\!\leq \!\E\left[\sum_{s=0}^{n-1}\!\left(\!\frac{1}{G_{1s}}\!-\! 1\!\right)\!\right]\! +\! \E\left[\sum_{s=0}^{n-1}\!\I\left\{\!G_{is}\!>\!\frac{1}{n}\right\}\right]+1.\! \! \end{equation}
\end{lemma}
Similar to the standard MAB formulation, the first term can be controlled  as the Gamma distribution has a heavy tail. Specifically,
\begin{equation*}
	  \frac{1}{G_{1s}} - 1 \geq   \frac{2 }{\prob(\widehat{\sigma}^2_{i,s}\geq\sigma_1^2, \widehat{\mu}_{i,s} \leq \mu_1)}-1.
\end{equation*}
Hence, to bound the first term in $(\ref{equ: equ15})$, we need a lower bound on the tail of the distribution of the empirical mean-variance.  Note that given $T_{1,t} = s$, the random variables $ \widehat{\mu}_{1,s} = \mu, \widehat{\sigma}_{1,s}^2 = \sigma^2$, $\theta_{1,t}$ and $\tau_{1,t}$ are independent because we sample them from different distributions independently. 

\begin{lemma}[Tail Lower Bound]
\label{lemma: lem3}
We have {\em 
\begin{align}
&\prob\big(\widehat{\MV}_{1,t}\! \geq \! \MV_1 -(1\! +\! \rho)\varepsilon \,\big|\, T_{1,t}\! =\! s,\widehat{\mu}_{1,s}\! =\! \mu,\widehat{\sigma}_{1,s}^2 \! =\!  \sigma^2\big) \nonumber\\
&\ge\left\{  \begin{array}{ll}
\prob\left(\frac{1}{\tau_{1,t}}-\sigma_1^2\leq \varepsilon\right)\cdot\prob\left(\theta_{1,t}-\mu_1\geq  - \varepsilon\right) &  \\
& \hspace{-.7in}\mbox{if  }\; \sigma^2\geq\sigma_1^2, \mu \leq \mu_1 \\
\frac{1}{2}\prob\left(\frac{1}{\tau_{1,t}}-\sigma_1^2\leq \varepsilon\right) & \hspace{-.7in} \mbox{if  }\; \sigma^2\geq\sigma_1^2, \mu > \mu_1 \\
\frac{1}{2}\prob\left(\theta_{1,t}-\mu_1\geq -\varepsilon\right)  & \hspace{-.7in}\mbox{if  }\; \sigma^2 < \sigma_1^2, \mu \le \mu_1 \\
\frac{1}{4}  
& \hspace{-.7in}\mbox{if  } \; \sigma^2 < \sigma_1^2, \mu > \mu_1 
\end{array}  \right. . \notag 
\end{align}}
\end{lemma}	

Lemma~\ref{lemma: lem3} helps us to split the lower bounding into two parts. Now we can deal with mean and variance separately. The tail probability bound of Gaussian distribution is standard, but for Gamma distribution, we devise a new and non-standard anti-concentration bound that is crucial in the analysis of Thompson Sampling for mean-variance MABs. 

\begin{lemma} \label{lemma: lem4}
	For a Gamma random variable $X\sim \mathrm{Gamma}\left(\alpha, \beta\right)$ with shape $\alpha \geq 2$ and rate $\beta>0$, 
\begin{equation}	
	\mathbb{P}\left(X\geq x\right)\geq \frac{1}{\Gamma(\alpha)}\exp\left(-\beta x\right)\left(1+\beta x\right)^{\alpha-1}, \;  \mathrm{for }\ x>0. \nonumber
\end{equation}	
\end{lemma}
We obtain an upper bound of the first term in $(\ref{equ: equ15})$ by plugging the bound in Lemma~\ref{lemma: lem4} into the terms in Lemma~\ref{lemma: lem3} and integrating the conditional probability over $(\mu,\sigma^2)$.

\begin{figure*}[t]
   \begin{minipage}{0.33\textwidth}
     \centering
     \includegraphics[width = 1\linewidth]{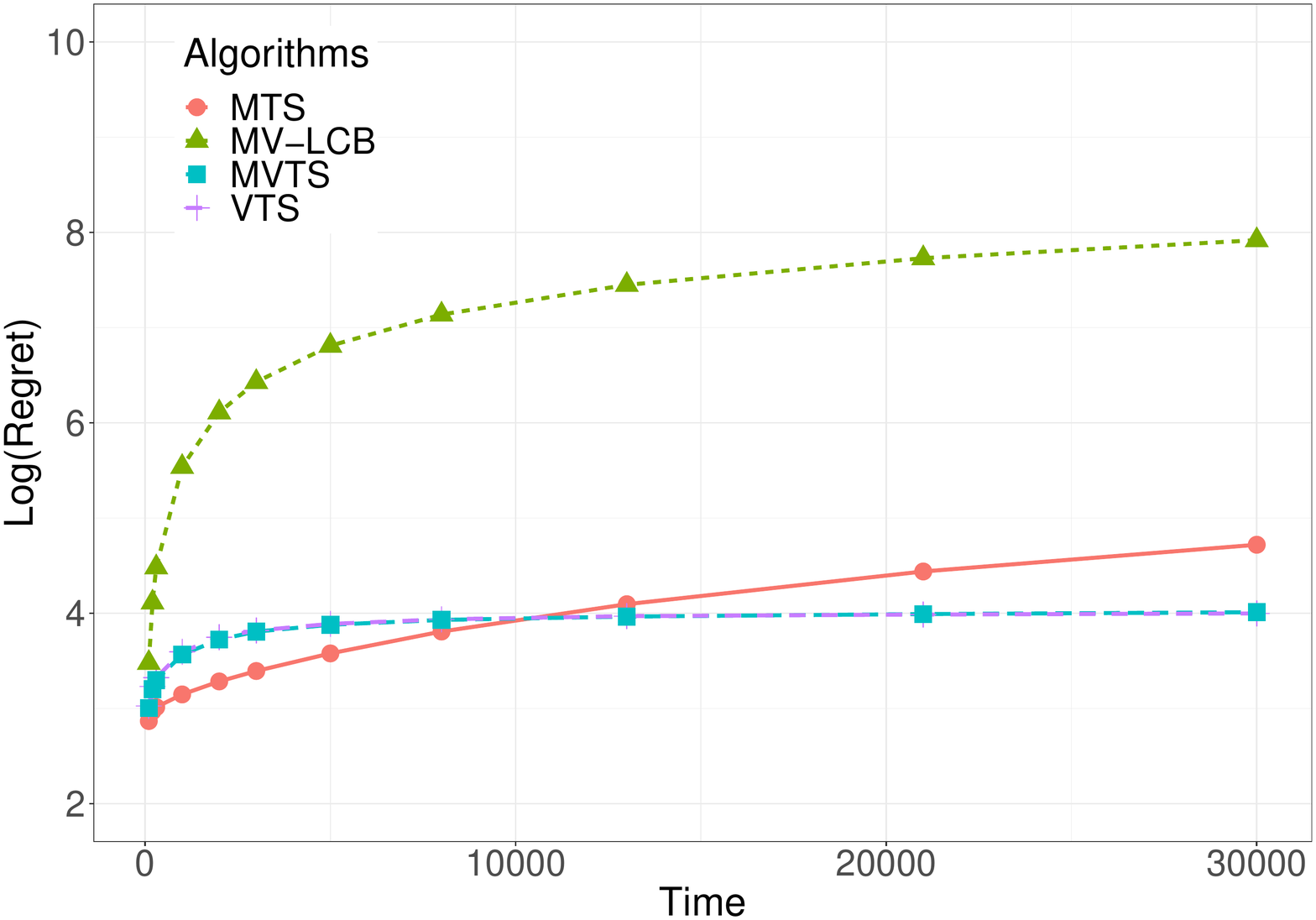}
     \caption{Regrets for $\rho = 10^{-3}$}\label{fig: fig2}
   \end{minipage}\hfill
   \begin{minipage}{0.33\textwidth}
     \centering
     \includegraphics[width = 1\linewidth]{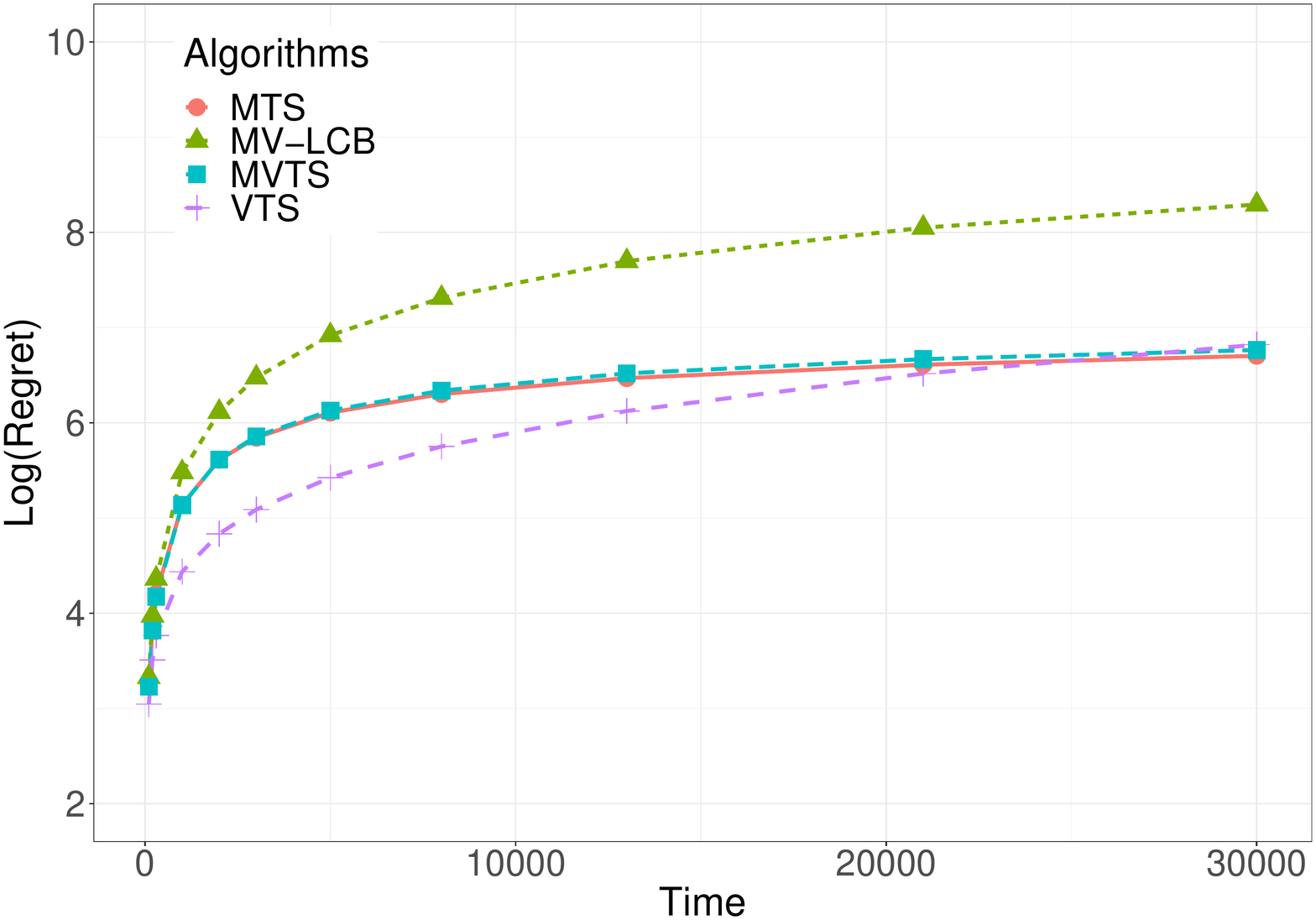}
     \caption{Regrets for $\rho = 1$}\label{fig: fig3}
   \end{minipage}
   \begin{minipage}{0.33\textwidth}
     \centering
     \includegraphics[width = 1\linewidth]{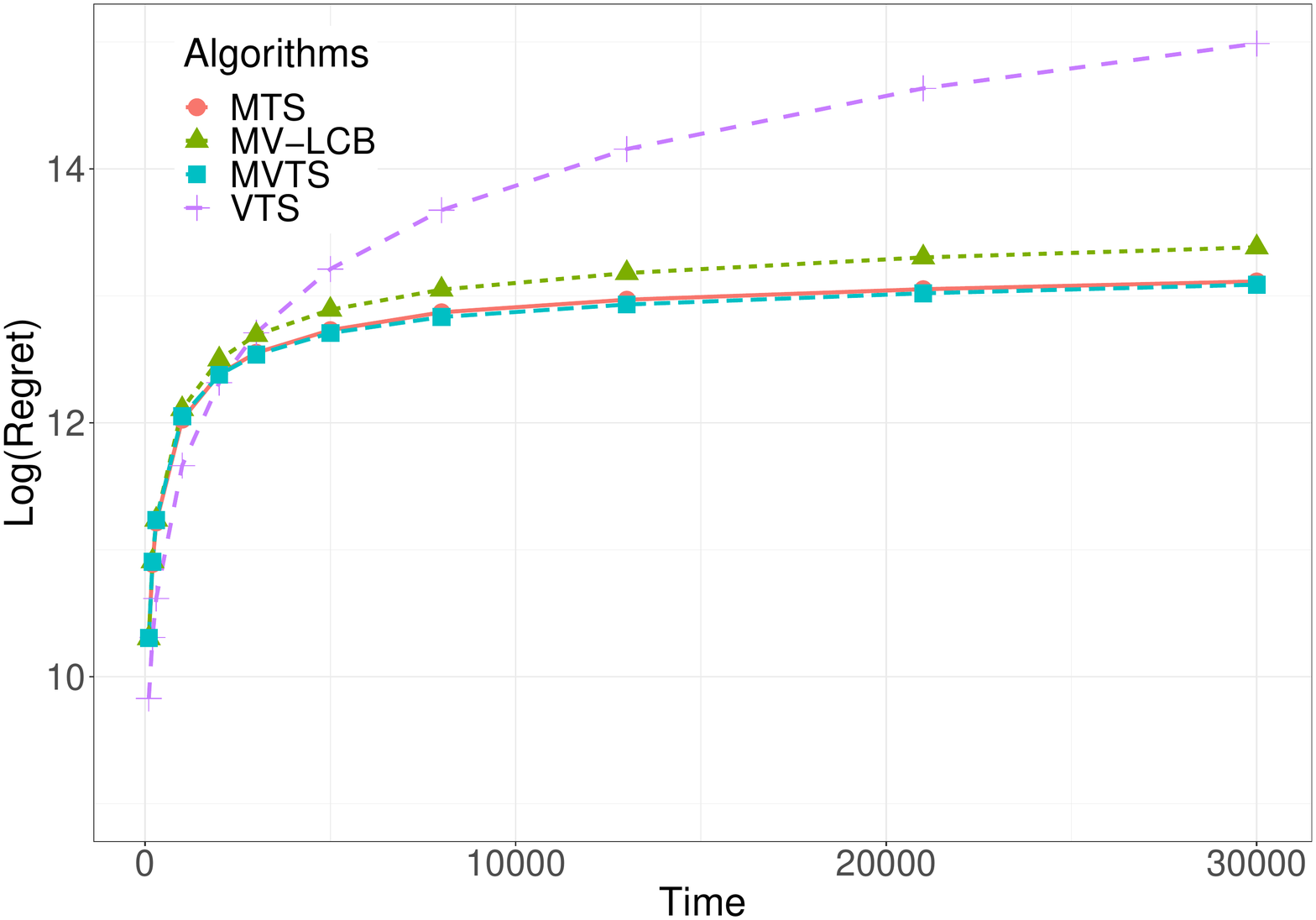}
     \caption{Regrets for $\rho = 1000$}\label{fig: fig4}
   \end{minipage}
\end{figure*}
\begin{figure*}[t]
   \begin{minipage}{0.48\textwidth}
     \centering
     \includegraphics[width = 6.9cm]{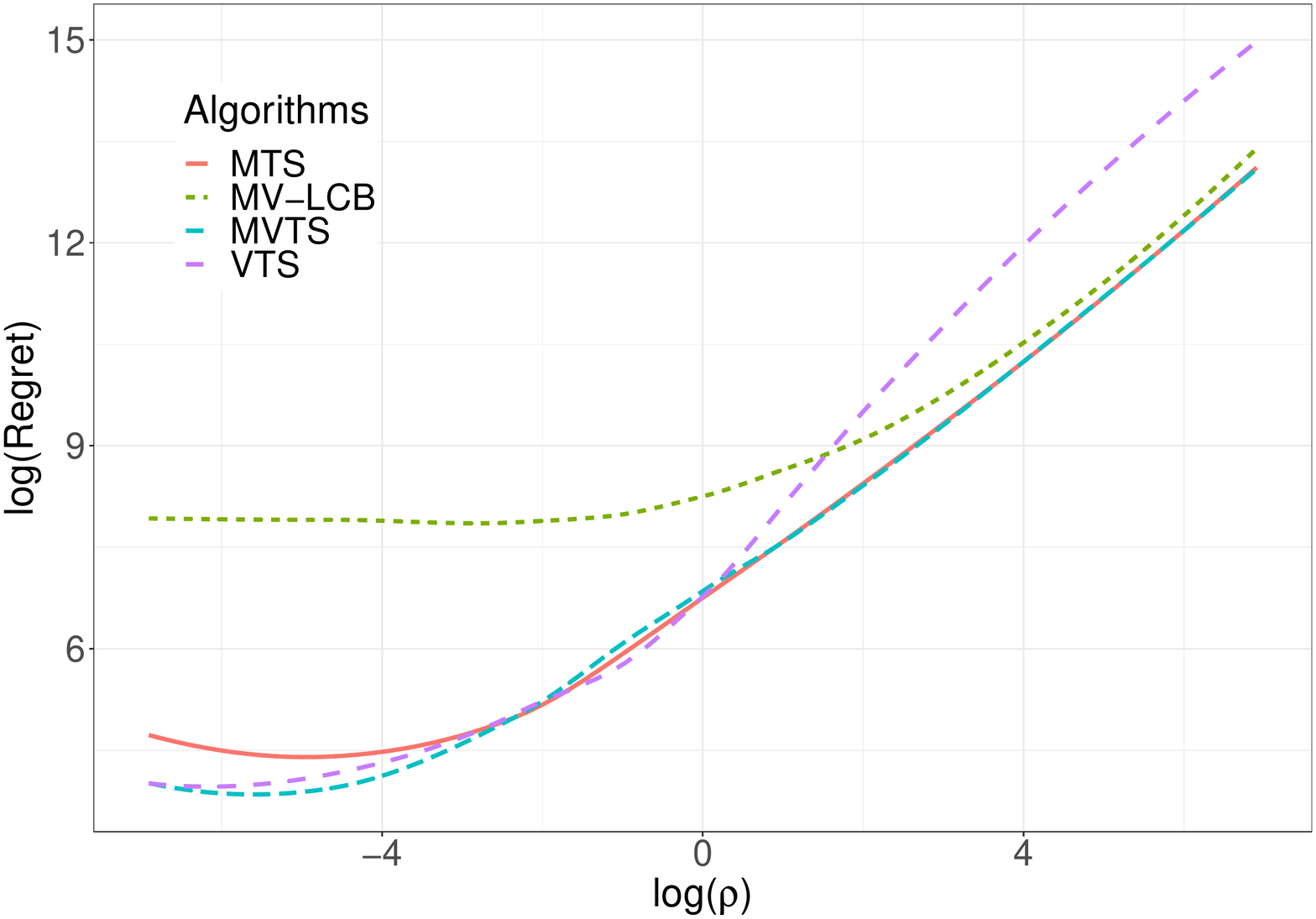}
     \caption{Regret of Gaussian MV MAB with $K=15$.}\label{fig: fig5}
   \end{minipage}\hfill
   \begin{minipage}{0.48\textwidth}
     \centering
     \includegraphics[width = 6.9cm]{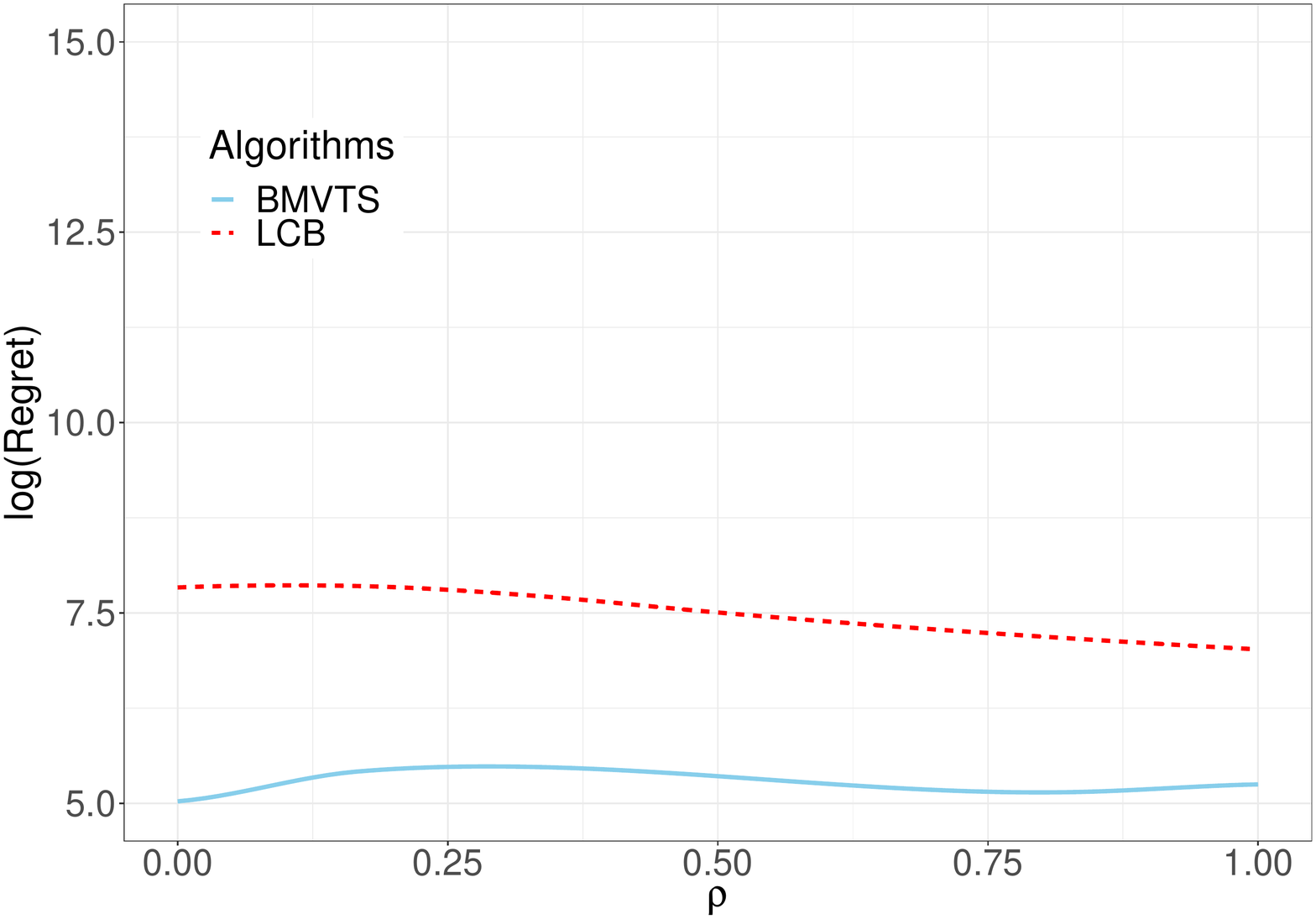}
     \caption{Regret of Bernoulli MV MAB with $K=15$..} \label{fig: fig6}
   \end{minipage}
\end{figure*}

For the second term, essentially, we need to bound $\prob\left(G_{is}>1/n\right)$. 
By direct computation, 
\begingroup
\allowdisplaybreaks
\begin{align*}
	&\prob\big(\widehat{\MV}_{i,s} \geq \MV_1 -(1+\rho)\varepsilon\,\big|\,T_{i,t} \!=\! s,\widehat{\mu}_{i,s}\!=\!\mu,\widehat{\sigma}_{i,s}^2 \!=\! \sigma^2\big)\\  
	&=  \prob\left(\rho\theta_{i,t} - \frac{1}{\tau_{i,t}}\geq \rho\mu_1 - \sigma_1^2-(1+\rho)\varepsilon\right) \\
	&= \prob\left(\rho(\theta_{i,t}-\mu_1) + \left(\sigma_1^2-\frac{1}{\tau_{i,t}}\right)\geq -(1+\rho)\varepsilon\right)\\
    &\leq  \prob\left(\theta_{i,t}-\mu_1\geq -\varepsilon\right) + \prob\left(\frac{1}{\tau_{i,t}} - \sigma_1^2\leq \varepsilon\right).
\end{align*}%
\endgroup
Hence, we have following key relationship
\begin{equation}
	\left\{G_{is}\!>\!\frac{1}{n}\right\} \;\Longrightarrow\;\begin{cases}
\prob\left(\tau_{i}(t)\geq \frac{1}{\sigma_i^2+\varepsilon}\right)\!\geq \!\frac{1}{2n}\\
\mathrm{ or }\\
\prob\left(\theta_{i,t}-\mu_1\geq -\varepsilon\right)\! \geq \!\frac{1}{2n}	
 \end{cases} \label{eqn:relat} .
\end{equation}
This relation presents a method to bound the probability of  \[\left\{G_{is}\! >\!\frac{1}{n}\right\}\!=\! \left\{\prob\left(\rho\theta_{i,t}\!-\!\frac{1}{\tau_{i,t}}\!\geq\! \MV_1\!-\!(1\!+\!\rho)\varepsilon\right)\!>\!\frac{1}{n}\right\}.\] The probability of this event is {\em a priori} not straightforward to bound because $\theta_{i,t}$ and $\tau_{i,t}$ are random variables and so are the parameters that define them (cf.\ Figure~\ref{fig: structure}). However, equipped with~\eqref{eqn:relat}, we can {decouple the Thompson samples representing the mean and variance}. To bound the second term on right of~\eqref{equ: equ15}, we use an upper bound on the tail of the Gamma distribution.\begin{lemma}[\citet{harremoes2016bounds}]\label{lemma: lem5}
Under the conditions of Lemma~\ref{lemma: lem4},
	\begin{equation}\mathbb{P}\left(X\geq x\right) \leq \exp\left(-2\alpha h\Big(\frac{\beta x}{\alpha}\Big)\right), \quad \mathrm{for }\ x>\frac{\alpha}{\beta}.
	\end{equation}
\end{lemma}
Plugging the bounds in  Lemmas~\ref{lemma: lem3} and~\ref{lemma: lem4} into~\eqref{equ: equ15} in   Lemma~\ref{lemma: lem2} allows us to bound the regret of MVTS.  We present complete proofs of the regrets of MTS, VTS and MVTS in the supplementary material.

\section{Numerical Simulations}
\label{sec: simulation}
There are   other algorithms that achieve the optimal regret bound, such as {MV-LCB}  \citep{vakili2016risk}. Due to the complexity of the problem and the differences in the assumptions on the reward distributions (e.g., MV-LCB in \citet{vakili2016risk} assumes the variances of the arm distributions are sub-Gaussian), it is difficult to perform a fair comparison of their theoretical regret bounds. {We emphasize though that our analyses require less stringent assumptions.} Hence, we compare these algorithms via extensive numerical simulations in this section. {The R code for all our experiments is provided along with this submission.}

We report numerical simulations to validate our theoretical results in the previous sections. We consider the variance minimization problem ($\rho =10^{-3}$),  the expected reward maximization problem  ($\rho = 1000$), and  an intermediate case ($\rho=1$). The $K=15$ Gaussian arms are set to the same as the experiments from \citet{sani2012risk} (i.e. $\mu = (0.1,0.2,0.23,0.27,0.32,0.32,0.34,0.41,0.43,0.54,0.55,$\\$0.56,0.67,0.71,0.79)$, $\sigma_i^2 = (0.05,0.34,0.28,0.09,0.23,$\\$0.72,0.19,0.14,0.44,0.53,0.24,0.36,0.56,0.49,0.85)$). We run MV-LCB, MTS, VTS and {MVTS}. 

In Figures~\ref{fig: fig2},~\ref{fig: fig3} and~\ref{fig: fig4} we present the expected regret $\mathcal{R}_n(\pi)$, which is averaged over $500$ runs. {The standard deviations of the regrets are small compared to the averages, and therefore are omitted from the all plots.} For $\rho =10^{-3}$, VTS outperforms all the other algorithms. For $\rho= 1000$, MTS has the  smallest regret compared to all the other algorithms. For $\rho = 1$, MTS, VTS and MVTS have similar performances, all of which are  better than MV-LCB. 

In order to validate our algorithms further and {to observe how they perform as functions of  $\rho$}, we ran our algorithms with different $\rho\in[10^{-3},1000]$. The time horizon $n =30,000$ is fixed and the regret is averaged over $500$ runs. We report the regrets in Fig.~\ref{fig: fig5}.  As expected, the regret of {MVTS} coincides with {VTS} when $\rho$ is small and with {MTS} when $\rho$ is large. We see also a significant performance improvement of {MVTS} compared to MV-LCB for all  $\rho\in \mathbb{R}_+$. 
 

The experiments on Bernoulli mean-variance bandits are presented in Fig.~\ref{fig: fig6}. Here the arm distributions are Bernoulli distributions with success probabilities $(0.1,0.2,0.23,0.27,0.32,0.32,0.34,0.41,0.43,0.54,0.55,$\\$0.56, 0.67,0.71,0.79)$. The regret is averaged over $500$ runs with a fixed time horizon $n = 30,000$. {We also designed and implemented an LCB-based algorithm (analogous to those designed by \citet{sani2012risk} and \citet{vakili2016risk}) for Bernoulli mean-variance bandits. However, Fig.~\ref{fig: fig3} clearly show that BMVTS significantly outperforms the LCB-based algorithm over all $\rho \in (0,1)$.}

\section{Conclusion}\label{sec:conc}
 To the best of our knowledge, this is the first work applying Thompson sampling to  solve risk-averse MAB problems.  We proved regret bounds that are asymptotically tight in certain  regimes and recover known results in other regimes. Experimental results show that our algorithms, particularly MVTS  beats the state-of-the-art LCB-style MAB algorithms for Bernoulli and Gaussian mean-variance bandits over all risk tolerance parameters~$\rho$. 

There are many different methods to model the risk-return trade-off, such as  CVaR (\citet{kolla2019risk}, \citet{xu2018index}.  \citet{galichet2013exploration}) and \textit{entropy risk} (\citet{maillard2013robust}) among others. Hence, more work is needed to explore the performance of Thompson Sampling, or indeed other algorithms, using different risk measures. We leave the regret analyses  for other risk measures and comparisons to the methodologies herein for future work.
 \section*{Acknowledgements}
We would like to express our gratitude to
the anonymous reviewers for their valuable comments which have helped to improve the presentation of this paper.   We also thank Joel Chang for pointing out some typos. 

This work is supported in part by a Singapore National Research Foundation (NRF) Fellowship under grant number R-263-000-D02-281.

\bibliography{citation}
\bibliographystyle{icml2020}

\newpage
\appendix
\onecolumn

\renewcommand{\theequation}{S-\arabic{equation}}
\renewcommand{\thesection}{S-\arabic{section}}
\renewcommand{\thefigure}{S-\arabic{figure}}
\renewcommand{\thetable}{S-\arabic{table}}
\renewcommand{\thetheorem}{S-\arabic{theorem}}
\renewcommand{\thealgorithm}{S-\arabic{algorithm}}
\renewcommand{\thelemma}{S-\arabic{lemma}}
\setcounter{lemma}{0}
\setcounter{theorem}{0}
\setcounter{figure}{0}
\setcounter{equation}{0}

\textbf{\Large Supplementary Material}
\section{Proof of Lemma \ref{lemma: lem1}}
\label{sec: S-1}
We first derive an upper bound for the expected cumulative regret. We use $\pi(t)$ to denote the arm that is pulled in period $t$ and $T_{i, j}$ is the number of times that arm $i$ is pulled during first $j$ periods. The time horizon $n$ will be fixed in the following proof. Given the definition of the empirical mean-variance in~(\ref{equ: equ3}), we rewrite the empirical mean as follows,
\[\widehat{\mu}_n\left(\pi\right) = \frac{1}{n}\sum_{t=1}^nX_{\pi\left(t\right),t} = \frac{1}{n}\sum_{i=1}^K T_{i,n}\widehat{\mu}_{i,T_{i,n}},\qquad \mu_1 = \frac{1}{n}\sum_{i=1}^K T_{i,n}\mu_1\]
where $\widehat{\mu}_{i,T_{i,n}}=\frac{1}{T_{i,n}}\sum_{t=1}^{n}X_{\pi(t),t}\mathbf{1}_{\pi(t)=i}$.

Similarly, the variance term can be written as
\begin{align*}
	\widehat{\sigma}_n^2(\pi) & = \frac{1}{n}\sum_{t=1}^{n}\left(X_{\pi(t),t} - \widehat{\mu}_n(\pi)\right)^2 = \frac{1}{n}\sum_{i=1}^K T_{i,n}\widehat{\sigma}_{i,T_{i,n}}^2 + \frac{1}{n}\sum_{i=1}^K T_{i,n}\left(\widehat{\mu}_{i,T_{i,n}}-\widehat{\mu}_n(\pi)\right)^2.
\end{align*}
We can further bound the second term as follows,
\[\frac{1}{n}\sum_{i=1}^K T_{i,n}\left(\widehat{\mu}_{i,T_{i,n}}-\widehat{\mu}_n(\pi)\right)^2 \leq \frac{1}{n}\sum_{i=1}^K\sum_{j\neq i} T_{i,n}T_{j,n}\left(\widehat{\mu}_{i,T_{i,n}}-\widehat{\mu}_{j,T_{j,n}}\right)^2.\]
Then 
\[\mathcal{R}_n(\pi) \leq \sum_{i=2}^K T_{i,n}\left(\MV_1-\widehat{\MV}_{i,T_{i,n}}\right) + \frac{1}{n}\sum_{i=1}^K\sum_{j\neq i} T_{i,n}T_{j,n}\left(\widehat{\mu}_{i,T_{i,n}}-\widehat{\mu}_{j,T_{j,n}}\right)^2.\]
Taking the expectation of the right hand side, we obtain
\begin{align*}
	&\E\left[\sum_{i=2}^K T_{i,n}\left(\MV_1-\widehat{\MV}_{i,T_{i,n}}\right) + \frac{1}{n}\sum_{i=1}^K\sum_{j\neq i} T_{i,n}T_{j,n}\left(\widehat{\mu}_{i,T_{i,n}}-\widehat{\mu}_{j,T_{j,n}}\right)^2\right]\\
	&= \sum_{i=2}^K 
	\E\left[\E\left[T_{i,n}\left(\MV_1-\widehat{\MV}_{i,T_{i,n}}\right)\,\Big|\, T_{i,n}\right]\right] + \frac{1}{n}\sum_{i=1}^K\sum_{j\neq i} \E\left[\E\left[T_{i,n}T_{j,n}\left(\widehat{\mu}_{i,T_{i,n}}-\widehat{\mu}_{j,T_{j,n}}\right)^2\,\Big|\, T_{i,n}, T_{j,n}\right]\right]\\
	&=\sum_{i=2}^K \E\left[T_{i,n} \left(\rho\mu_1-\sigma_i^2\frac{T_{i,n}-1}{T_{i,n}}-\rho\mu_i+\sigma_1^2 \right)\right]+ \frac{1}{n}\sum_{i=1}^K\sum_{j\neq i} \E\left[T_{i,n}T_{j,n}\left(\Gamma_{i,j}^2+\frac{\sigma_i^2}{T_{i,n}}+\frac{\sigma_j^2}{T_{j,n}}\right)\right]\\
	&=\sum_{i=2}^K\E\left[T_{i,n}\right]\Delta_i + \sum_{i=2}^K \sigma_i^2 + \frac{1}{n}\sum_{i=1}^K\sum_{j\neq i} \E\left[T_{i,n}T_{j,n}\Gamma_{i,j}^2\right] + \frac{1}{n}\sum_{i=1}^K\sum_{j\neq i} \E\left[\sigma_i^2T_{j,n} + \sigma_j^2T_{i,n}\right] \\ 
	&=\sum_{i=2}^K\E\left[T_{i,n}\right]\Delta_i + \frac{1}{n}\sum_{i=1}^K\sum_{j\neq i} \E\left[T_{i,n}T_{j,n}\right]\Gamma_{i,j}^2 + 3\sum_{i=1}^K \sigma_i^2 - \frac{2}{n}\sum_{i=1}^K \sigma_i^2\E\left[T_{i,n}\right] \\
	&\leq   \sum_{i=2}^K\E\left[T_{i,n}\right]\Delta_i + \frac{1}{n}\sum_{i=1}^K\sum_{j\neq i} \E\left[T_{i,n}T_{j,n}\right]\Gamma_{i,j}^2 + 3\sum_{i=1}^K \sigma_i^2.
\end{align*}
This completes the proof of Lemma~\ref{lemma: lem1}.
\section{Figures and more numerical results}
%


In this section, we show more numerical results to validate our theoretical results in the main paper.

\begin{figure*}[t]
   \begin{minipage}{0.33\textwidth}
     \centering
     \includegraphics[width = 1\linewidth]{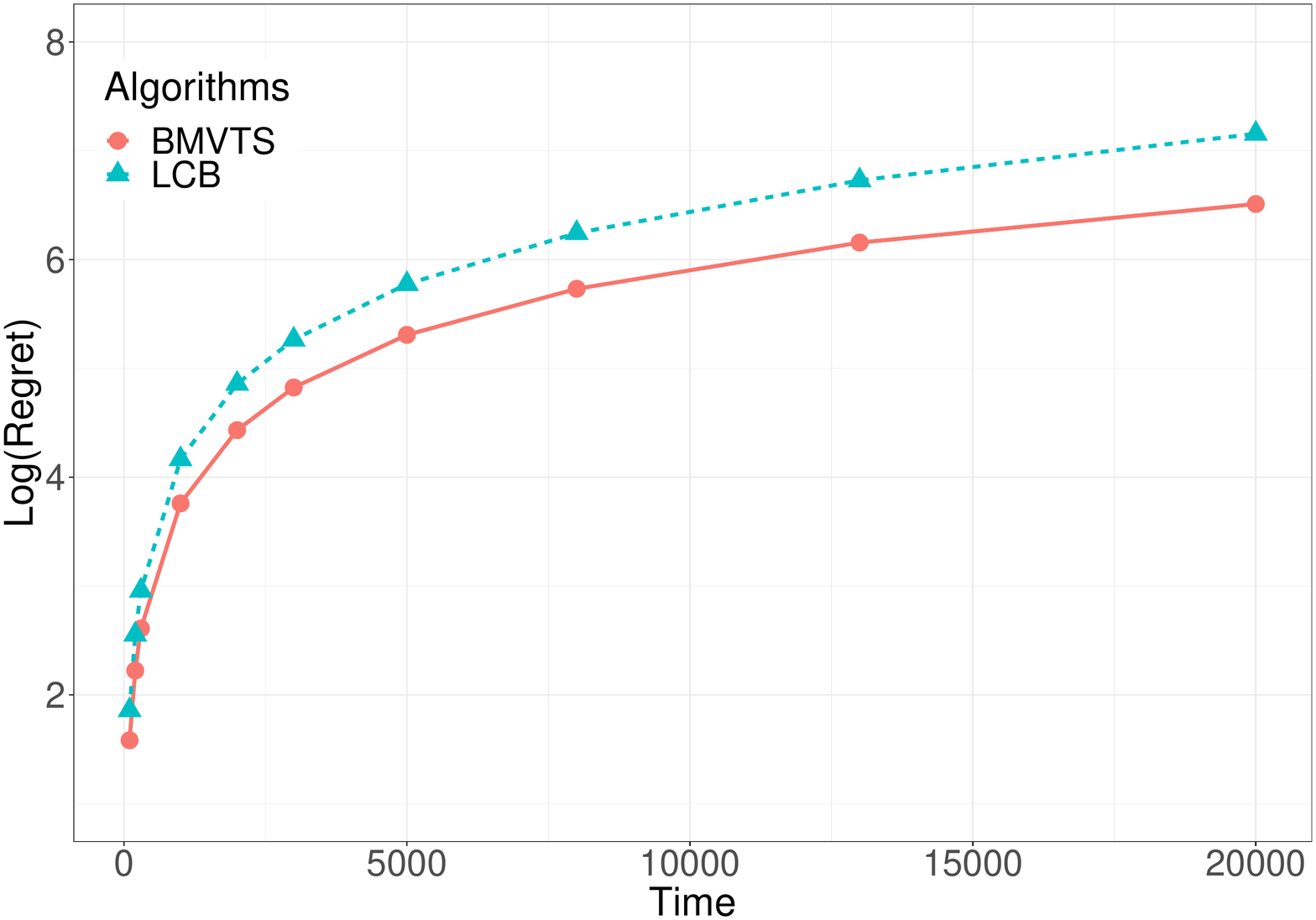}
     \caption{Regrets for $\rho = 0.111$}\label{fig: S-5}
   \end{minipage}\hfill
   \begin{minipage}{0.33\textwidth}
     \centering
     \includegraphics[width = 1\linewidth]{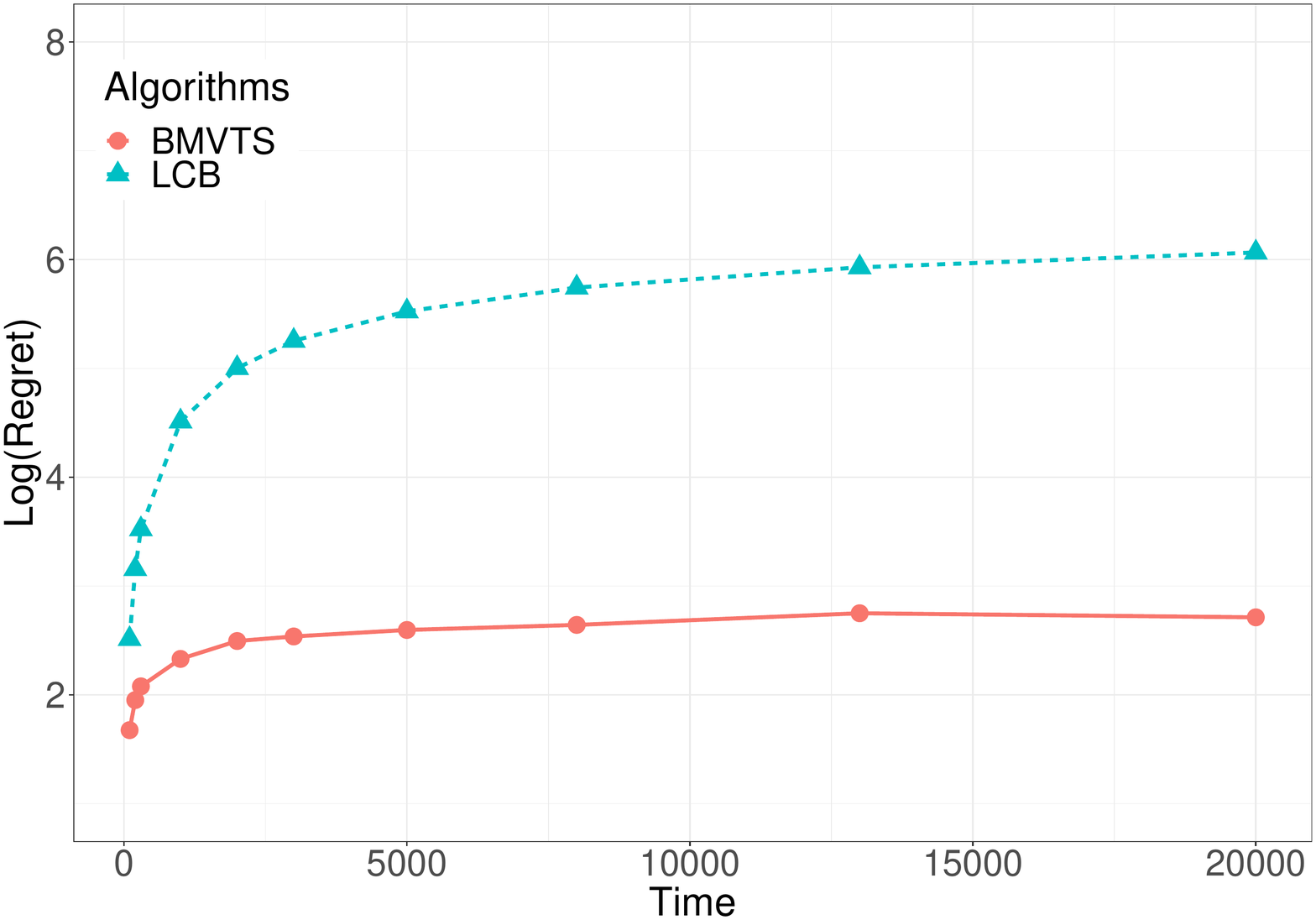}
     \caption{Regrets for $\rho = 0.444$}\label{fig: S-6}
   \end{minipage}
   \begin{minipage}{0.33\textwidth}
     \centering
     \includegraphics[width = 1\linewidth]{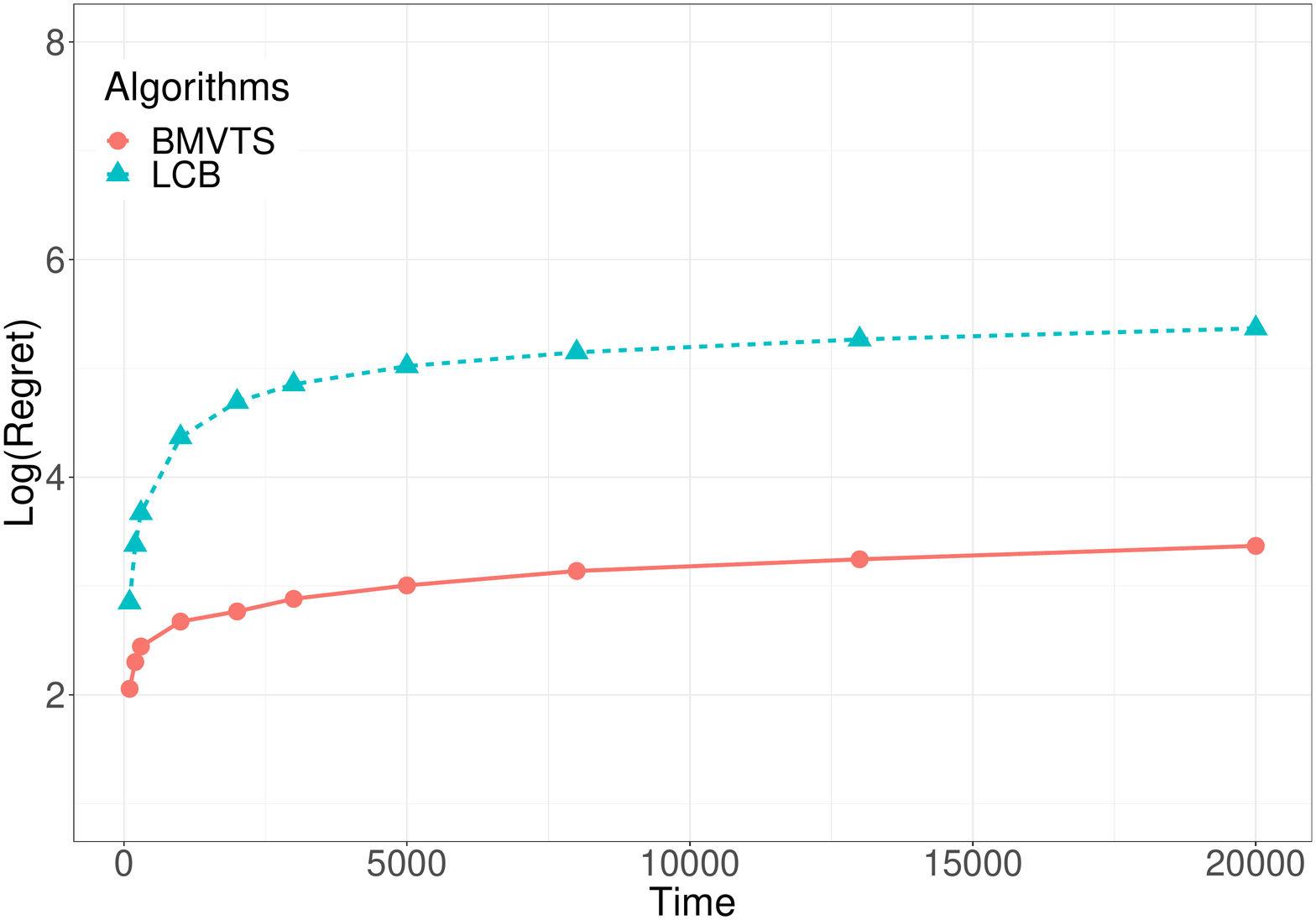}
     \caption{Regrets for $\rho = 0.889$}\label{fig: S-7}
   \end{minipage}
\end{figure*}

In Figures~\ref{fig: S-5},~\ref{fig: S-6} and~\ref{fig: S-7}, we report the expected regret of BMVTS with different $\rho$. Here the arm distributions are Bernoulli's with success probabilities $(0.1,0.2,0.23,0.27,0.32,0.32,0.34,0.41,0.43,0.54,0.55,0.56,0.67,0.71,0.79)$. The regret is averaged over $500$ runs with a fixed time horizon $n = 30000$. These figures clearly show that BMVTS outperform LCB algorithm. 

\section{Proof of Theorem~\ref{thm: thm3}}
\label{sec: S-2}
Since Theorem~\ref{thm: thm3} is the most involved, we present it before Theorems \ref{thm: thm1} and \ref{thm: thm2}. The proofs of the latter two theorems reuse several calculations that are done for the proof of Theorem~\ref{thm: thm3}.
\subsection{Notations}

\label{sec: S-2-1}
We remind the reader of the definitions of the event $E_i(t)$ and the probability $G_{is}$ as follows:
\begin{equation}E_i\left(t\right)=\left\{\widehat{\MV}_{i,t}\leq \MV_1-(1+\rho)\varepsilon\right\}, \quad G_{is} = \prob\left(E_i(t)^c|T_{i,t}=s\right).\nonumber\end{equation}

According to Lemma~\ref{lemma: lem2}, we need to provide an upper bound for $\E[T_{i,n}]$.

\subsection{Proofs of the lemmas}
\label{sec: S-2-2}
Before we get into the details of the proofs, let us present the proof of the lemmas in the main text and some other useful lemmas.
\begin{lemma}[Lemma~\ref{lemma: lem4} in the main text]
\label{lem: S-1}
	For a Gamma random variable $X\sim \mathrm{Gamma}\left(\alpha, \beta\right)$ with shape $\alpha \geq 2$ and rate $\beta>0$, we have the following lower bound on the complementary cumulative distribution function
\begin{equation}	
	\mathbb{P}\left(X\geq x\right)\geq \frac{1}{\Gamma(\alpha)}\exp\left(-\beta x\right)\left(1+\beta x\right)^{\alpha-1}, \;  \mathrm{for }\ x>0.\nonumber
\end{equation}	
\end{lemma}
\textit{Proof}: Let $Y$ be an exponential random variable with rate $\beta$.  Consider, 
\begin{align}
	\mathbb{P}\left(X\geq x\right) &= \int_{x}^\infty \frac{\beta^{\alpha} t^{\alpha-1} e^{-\beta t}}{\Gamma(\alpha)}\,\mathrm{d}t \nonumber \\
	& = \frac{\beta^\alpha e^{-\beta x}}{\Gamma(\alpha)} \int_{0}^{\infty}(z+x)^{\alpha-1} e^{-\beta z}\, \mathrm{d} z\nonumber \\
	& = \frac{\beta^{\alpha-1} e^{-\beta x}}{\Gamma(\alpha)} \int_{0}^{\infty}(z+x)^{\alpha-1} \beta e^{-\beta z}\, \mathrm{d}z\nonumber  \\
	& = \frac{\beta^{\alpha-1} e^{-\beta x}}{\Gamma(\alpha)} \mathbb{E}\big(\left[Y + x\right]^{\alpha-1}\big)\nonumber \\
	& \geq \frac{\beta^{\alpha-1} e^{-\beta x}}{\Gamma(\alpha)} \big(\mathbb{E}[Y + x]\big)^{\alpha-1} \label{eqn:jens}\\
	& = \frac{\beta^{\alpha-1} e^{-\beta x}}{\Gamma(\alpha)} \bigg(\frac{1}{\beta}+x\bigg)^{\alpha-1} \nonumber \\
	& = \frac{e^{-\beta x}}{\Gamma(\alpha)} \left(1+\beta x\right)^{\alpha-1}\nonumber 
\end{align}
where~\eqref{eqn:jens} follows from Jensen's inequality and the convexity of $z\mapsto z^{\alpha-1}$ (recall that $\alpha\ge 2$).

\begin{lemma}[\citet{harremoes2016bounds}]
\label{lem: S-2}
Under the same setting as Lemma~\ref{lem: S-1},
	\begin{equation}\mathbb{P}\left(X\geq x\right) \leq \exp\left(-2\alpha h\left(\frac{\beta x}{\alpha}\right)\right), \quad \mathrm{for }\ x>\frac{\alpha}{\beta}. \nonumber
	\end{equation}
	where $h(x) = (x - 1 -\log x) /2$.
\end{lemma}
In the proofs of the following two lemmas, we use the following fact: Given $T_{i,t} = s, \widehat{\mu}_{i,s} = \mu<\mu_1, \widehat{\sigma}^2_{i,s} = \sigma^2 > \sigma_1^2$, $\theta_{i,t}$ and $\tau_{i,t}$ are independent because we sample them from different distributions independently. 
\begin{lemma}[Tail Upper Bound]
\label{lem: S-3}
We have
    \[\prob\left(\widehat{\MV}_{i,s} \geq \MV_1 - (1+\rho)\varepsilon\,\Big|\, T_{i,t} = s,\widehat{\mu}_{i,s}=\mu,\widehat{\sigma}_{i,s}^2 = \sigma^2 \right)\leq \exp\left(-\frac{s}{2}\left(\mu_1-\mu-\varepsilon\right)^2\right) + \exp\left(-sh\left(\frac{\sigma^2}{\sigma_1^2+ \varepsilon}\right)\right).\]
\end{lemma}
\textit{Proof:}
We can compute this probability directly, 
\begin{align*}
	& \prob(\widehat{\MV}_{i,s} \geq \MV_1 -(1+\rho)\varepsilon\,\big|\,T_{i,t} = s,\widehat{\mu}_{i,s}=\mu,\widehat{\sigma}_{i,s}^2 = \sigma^2)  \\
	&= \prob\left(\rho\theta_{i,t} - \frac{1}{\tau_{i,t}}\geq \rho\mu_1 - \sigma_1^2- (1+\rho)\varepsilon\right) \\
	&=\prob\left(\rho(\theta_{i,t}-\mu_1) + \left(\sigma_1^2-\frac{1}{\tau_{i,t}}\right)\geq -(1+\rho)\varepsilon\right)\\
	&\le  \prob\left(\theta_{i,t}-\mu_1\geq -\varepsilon\right) + \prob\left(\frac{1}{\tau_{i,t}} - \sigma_1^2\leq \varepsilon\right)\\
	&\le  \exp\left(-\frac{s}{2}\left(\mu_1-\mu-\varepsilon\right)^2\right) + \prob\left(\tau_{i,t}\geq \frac{1}{\sigma_1^2+ \varepsilon}\right) \\
	&\le \exp\left(-\frac{s}{2}\left(\mu_1-\mu-\varepsilon\right)^2\right) + \exp\left(-sh\left(\frac{\sigma^2}{\sigma_1^2+ \varepsilon}\right)\right).
\end{align*}
The last inequality follows from Lemma~\ref{lem: S-2}.
The bound  in Lemma~\ref{lem: S-3} is crucial for proving an upper bound of $G_{is}$, which is presented in Section~\ref{sec: S-2-4}.

\begin{lemma}[Lemma~\ref{lemma: lem3} in the main text]
\label{lem: S-4}
We have 
{\em 
\begin{align}
&\prob\big(\widehat{\MV}_{1,t}\! \geq \! \MV_1 -(1\! +\! \rho)\varepsilon \,\big|\, T_{1,t}\! =\! s,\widehat{\mu}_{1,s}\! =\! \mu,\widehat{\sigma}_{1,s}^2 \! =\!  \sigma^2\big) \nonumber\\
&\ge\left\{  \begin{array}{ll}
\prob\left(\frac{1}{\tau_{1,t}}-\sigma_1^2\leq \varepsilon\right)\prob\left(\theta_{1,t}-\mu_1\geq  - \varepsilon\right) &  \\
& \hspace{-.7in}\mbox{if  }\; \sigma^2\geq\sigma_1^2, \mu \leq \mu_1 \\
\frac{1}{2}\prob\left(\frac{1}{\tau_{1,t}}-\sigma_1^2\leq \varepsilon\right) & \hspace{-.7in} \mbox{if  }\; \sigma^2\geq\sigma_1^2, \mu > \mu_1 \\
\frac{1}{2}\prob\left(\theta_{1,t}-\mu_1\geq -\varepsilon\right)  & \hspace{-.7in}\mbox{if  }\; \sigma^2 < \sigma_1^2, \mu \le \mu_1 \\
\frac{1}{4}  
& \hspace{-.7in}\mbox{if  } \; \sigma^2 < \sigma_1^2, \mu > \mu_1 
\end{array}  \right. . \label{equ: S-2}
\end{align} }
\end{lemma}
\textit{Proof:} Consider the following set of equalities and inequality,
\begin{align}
	&\prob \left(\widehat{\MV}_{1,t} \geq \MV_1 -(1+\rho)\varepsilon\,\big|\,T_{1,t} = s,\widehat{\mu}_{1,s}=\mu,\widehat{\sigma}_{1,s}^2 = \sigma^2\right)\nonumber  \\
	&=\prob\left(\rho\theta_{1,t} - \frac{1}{\tau_{1,t}}\geq \rho\mu_1 - \sigma_1^2-(1+\rho)\varepsilon\right) \nonumber \\
    &= \prob\left(\rho(\theta_{1,t}-\mu_1) - \left(\frac{1}{\tau_{1,t}}-\sigma_1^2\right)\geq -(1+\rho)\varepsilon\right)\nonumber \\
	&\ge \prob\left(\theta_{1,t}-\mu_1\geq -\varepsilon\right) \cdot \prob\left(\frac{1}{\tau_{1,t}}-\sigma_1^2\leq \varepsilon\right).
	\label{equ: S-3}
\end{align}
Then the lemma is proved by the inequality in (\ref{equ: S-3}), and
\begin{equation}\label{equ: S-4}\prob\left(\theta_{1,t}-\mu_1\geq -\varepsilon\right) > \frac{1}{2} \quad \text{ if } \quad \mu >\mu_1\end{equation}
and
\begin{equation}\label{equ: S-5}\prob\left(\frac{1}{\tau_{1,t}}-\sigma_1^2\leq \varepsilon\right)\geq \frac{1}{2} \quad \text{ if } \quad \sigma^2<\sigma_1^2.\end{equation}
Note that (\ref{equ: S-4}) and (\ref{equ: S-5}) can be established by using the properties of the median of Gaussian and Gamma distributions respectively. 
\\ Lemma \ref{lem: S-4} provides  us with a lower bound on $G_{1s}$, which is useful when we prove an upper bound  of $\E\left[\frac{1}{G_{1s}}-1\right]$ in Section~\ref{sec: S-2-3}.
\subsection{Bounding the first term of (\ref{equ: equ15})}
\label{sec: S-2-3}
We now provide a bound for the first term of (\ref{equ: equ15}) in Lemma~\ref{lemma: lem2}.

Let $c_1 = 1/\sqrt{2\pi\sigma_1^2},c_2 = \frac{1}{2^{s/2} \Gamma\left(s/2\right)\sigma_1^{s-2}},\tau = s(\sigma_1^2+\varepsilon)$ and fix $\varepsilon>0$. We define the conditional version of $G_{1s}$ as  \[\widetilde{G}_{1s} = G_{1s} |_{\widehat{\mu}_{1,s} = \mu, \widehat{\sigma}_{1,s}^2 = \beta}= \prob\left(\widehat{\MV}_{i,t}\geq \MV_1-(1+\rho)\varepsilon\,\big|\,\widehat{\mu}_{1,s} = \mu, \widehat{\sigma}_{1,s}^2 = \beta\right)\] 
which is the left-hand-side of (\ref{equ: S-2}) in Lemma~\ref{lem: S-4}.  
Then we calculate the expectation $\E\left[\frac{1}{G_{1s}}-1\right]$ by conditioning on various values of $\widehat{\mu}_{i,s}$ and $\widehat{\sigma}_{i,s}^2$.
 Note that we assumed that $\sigma_i^2\leq 1$ for all $i = 1,\ldots,K$.
For clarity, we partition the parameter space $(\beta,\mu)\in[0,\infty)\times (-\infty, \infty) $ into four parts as follows
\begin{align*}
 [0,\infty]\times(-\infty,\infty) = A \cup B\cup C\cup D
\end{align*}
where 
\begin{align*}
 A &= [0,\tau)\times[\mu_1-\varepsilon,\infty), \quad\,\, B= [0,\tau)\times(-\infty,\mu_1-\varepsilon], \\
   C &= [\tau,\infty)\times[\mu_1-\varepsilon,\infty),\quad  D = [\tau,\infty)\times(-\infty,\mu_1-\varepsilon].
\end{align*}
Then the expectation of $(1/G_{1s})-1$ can be partitioned into four parts as follows, 
\begin{align}
	\E\left[\frac{1}{G_{1s}}-1\right] & = c_1c_2\int_{0}^\infty\int_{-\infty}^\infty \frac{1-\widetilde{G}_{1s}}{\widetilde{G}_{1s}} \exp\left(-\frac{s(\mu-\mu_1)^2}{2\sigma_1^2}\right)\beta^{\frac{s}{2}-1} e^{-\frac{\beta}{2\sigma_1^2}}\,\mathrm{d}\mu\,\mathrm{d}\beta \nonumber\\
	& = c_1c_2\left(\int_A+\int_B+\int_C+\int_D\right)\frac{1-\widetilde{G}_{1s}}{\widetilde{G}_{1s}} \exp\left(-\frac{s(\mu-\mu_1)^2}{2\sigma_1^2}\right)\beta^{\frac{s}{2}-1} e^{-\frac{\beta}{2\sigma_1^2}}\,\mathrm{d}\mu\,\mathrm{d}\beta \label{equ: S-6}
\end{align}
\\\textit{Part A:}
Using the fourth case in Lemma~\ref{lem: S-4}, we have
$$ \frac{ 1-\widetilde{G}_{1s}}{\widetilde{G}_{1s}}\geq 4(1-\widetilde{G}_{1s}).$$ Then
\begin{align}
	&\ c_1c_2\int_A \frac{1-\widetilde{G}_{1s}}{\widetilde{G}_{1s}} \exp\left(-\frac{s(\mu-\mu_1)^2}{2\sigma_1^2}\right)\beta^{\frac{s}{2}-1} e^{-\frac{\beta}{2\sigma_1^2}}\,\mathrm{d}\mu\,\mathrm{d}\beta\nonumber\\
	&\leq  4c_1c_2\int_A (1-\widetilde{G}_{1s})\exp\left(-\frac{s(\mu-\mu_1)^2}{2\sigma_1^2}\right)\beta^{\frac{s}{2}-1} e^{-\frac{\beta}{2\sigma_1^2}} \,\mathrm{d}\mu\,\mathrm{d}\beta\nonumber\\
&\le 4c_1c_2\int_A \left(\prob\left(\theta_{1,t}-\mu_1 \leq -\varepsilon \,\big|\,\widehat{\mu}_{1,s} = \mu\right)+\prob\left(\frac{1}{\tau_{1,t}} -\sigma_1^2 \geq \varepsilon\,\Big|\,\widehat{\sigma}_{1,s}^2 = \beta\right)\right) \nonumber\\
&\qquad\cdot\exp\left(-\frac{s(\mu-\mu_1)^2}{2\sigma_1^2}\right)\beta^{\frac{s}{2}-1} e^{-\frac{\beta}{2\sigma_1^2}}\,\mathrm{d}\mu\,\mathrm{d}\beta\nonumber\\
&\le 4c_1\int_{\mu_1}^{\infty} \prob\left(\theta_{1,t}-\mu_1 \leq -\varepsilon\,\big|\,\widehat{\mu}_{1,s} = \mu\right) \exp\left(-\frac{s(\mu-\mu_1)^2}{2\sigma_1^2}\right)\,\mathrm{d}\mu\nonumber\\
&\qquad  + 4c_2\int_{\mu_1}^{\infty} \prob\left(\frac{1}{\tau_{1,t}} -\sigma_1^2 \geq \varepsilon\,\Big|\,\widehat{\sigma}_{1,s}^2 = \beta\right) \beta^{\frac{s}{2}-1} e^{-\frac{\beta}{2\sigma_1^2}}\,\mathrm{d}\beta\nonumber\\
&\le  \ 4c_1\int_{\mu_1}^{\infty} \exp\left(-\frac{s(\mu-\mu_1+\varepsilon)^2}{2}\right) \exp\left(-\frac{s(\mu-\mu_1)^2}{2\sigma_1^2}\right)\,\mathrm{d}\mu  \nonumber \\
&\qquad+ 4c_2\int_{0}^\tau\exp\left(-\frac{\left(\beta-s\left(\sigma_1^2+\varepsilon\right)\right)^2}{4s\left(\sigma_1^2+\varepsilon\right)^2}\right)\beta^{\frac{s}{2}-1} e^{-\frac{\beta}{2\sigma_1^2}}\,\mathrm{d}\beta  \label{equ: S-7}\\
&\le  \dfrac{8c_1}{s\varepsilon}\exp\left(-\frac{s\varepsilon^2}{2}\right) + 4\exp\left(-\frac{s\varepsilon^2}{4(\sigma_1^2+\varepsilon)^2}\right) \\
&\le C_1\exp\left(-\frac{s\varepsilon^2}{4}\right) \nonumber
\end{align}
where (\ref{equ: S-7}) follows from using  tail upper bounds on the  Gaussian and Gamma distributions. Here, and in the following, we use the notation $C_i, i\in\mathbb{N}$ to denote constants. 
 \\\\\textit{Part B:} Using the third case in Lemma~\ref{lem: S-4}, we have $$\frac{ 1-\widetilde{G}_{1s} }{\widetilde{G}_{1s}}\geq\frac{ 2}{\prob\left(\theta_{1,t}-\mu_1\geq -\varepsilon\,\big|\,\widehat{\mu}_{1,s} = \mu\right)}.$$ Then
\begin{align}
	& \ c_1c_2\int_B \frac{1-\widetilde{G}_{1s}}{\widetilde{G}_{1s}} \exp\left(-\frac{s(\mu-\mu_1)^2}{2\sigma_1^2}\right)\beta^{\frac{s}{2}-1} e^{-\frac{\beta}{2\sigma_1^2}}\,\mathrm{d}\mu\,\mathrm{d}\beta\nonumber\\
	&\le  2c_1c_2\int_B \frac{1}{\prob\left(\theta_{1,t}-\mu_1\geq -\varepsilon\,\big|\,\widehat{\mu}_{1,s} = \mu\right)}\exp\left(-\frac{s(\mu-\mu_1)^2}{2\sigma_1^2}\right)\beta^{\frac{s}{2}-1} e^{-\frac{\beta}{2\sigma_1^2}}\,\mathrm{d}\mu\,\mathrm{d}\beta\nonumber\\
	&\le  2c_1\int_{-\infty}^{\mu_1-\varepsilon}\frac{1}{\prob\left(\theta_{1,t}-\mu_1\geq -\varepsilon\,\big|\,\widehat{\mu}_{1,s} = \mu\right)}\exp\left(-\frac{s(\mu-\mu_1)^2}{2\sigma_1^2}\right)\mathrm{d}\mu \cdot c_2\int_{0}^\tau \beta^{\frac{s}{2}-1} e^{-\frac{\beta}{2\sigma_1^2}}\mathrm{d}\beta \nonumber\\
	&\le  2c_1\int_{-\infty}^{\mu_1-\varepsilon}\frac{1}{\prob\left(\theta_{1,t}-\mu_1\geq -\varepsilon\,\big|\,\widehat{\mu}_{1,s} = \mu\right)}\exp\left(-\frac{s(\mu-\mu_1)^2}{2\sigma_1^2}\right)\mathrm{d}\mu \nonumber\\
	&\le  2c_1\int_{-\infty}^{\mu_1-\varepsilon} \left(\sqrt{s}(\mu_1-\mu-\varepsilon)+\sqrt{s(\mu_1-\mu-\varepsilon)^2+4}\right)\exp\left(\frac{s(\mu-\mu_1+\varepsilon)^2}{2}\right)\exp\left(-\frac{s(\mu-\mu_1)^2}{2\sigma_1^2}\right)\mathrm{d}\mu \label{equ: S-8}\\
	&\le 2 c_1\int_{0}^{\infty}\left(\sqrt{s}z+\sqrt{sz^2+4}\right)\exp\left(\frac{sz^2}{2} \right)\exp\left(-\frac{s(z+\varepsilon)^2}{2}\right)\mathrm{d}z \nonumber\\
	&\le 2c_1 \exp\left(-\frac{s\varepsilon^2}{2}\right)\int_{0}^\infty C_2\sqrt{s}z\exp\left(-sz\varepsilon\right)\mathrm{d}z\nonumber\\
	&\le C_3 \exp\left(-\frac{s\varepsilon^2}{2} \right)\label{equ: S-9}.
\end{align}
For (\ref{equ: S-8}), we use the following well-known lower bound for the tail of Gaussian distribution (see for example, Formula~7.1.13 in~\citet{Abr65}). Namely,   for a Gaussian random variable $X$ with mean $\mu$ and variance $\sigma^2$, we have 
\[\mathbb{P}(X\geq \mu+\sigma x)\geq \sqrt{\frac{2}{\pi}}\cdot\frac{1}{x+\sqrt{x^2+4}}\exp\left(-\frac{x^2}{2}\right) ,\quad\forall\,x\geq 0.\] For \eqref{equ: S-9}, we used integration by parts.
  \\\\\textit{Part C:} Use the second case in Lemma~\ref{lem: S-4}, we have $$ \frac{1-\widetilde{G}_{1s}}{\widetilde{G}_{1s}}\geq\frac{ 2}{\prob\left(\frac{1}{\tau_{1,t}}-\sigma_1^2\leq \varepsilon\,\big|\,\widehat{\sigma}_{i,s}=\beta\right)}.$$ Then, we have 
\begin{align}
	& c_1c_2\int_C \frac{1-\widetilde{G}_{1s}}{\widetilde{G}_{1s}} \exp\left(-\frac{s(\mu-\mu_1)^2}{2\sigma_1^2}\right)\beta^{\frac{s}{2}-1} e^{-\frac{\beta}{2\sigma_1^2}}\,\mathrm{d}\mu\,\mathrm{d}\beta\nonumber\\
	\leq &\ 2c_1c_2\int_C \frac{1}{\prob\left(\frac{1}{\tau_{1,t}}-\sigma_1^2\leq \varepsilon\,\big|\,\widehat{\sigma}_{i,s}=\beta\right)}\exp\left(-\frac{s(\mu-\mu_1)^2}{2\sigma_1^2}\right)\beta^{\frac{s}{2}-1} e^{-\frac{\beta}{2\sigma_1^2}}\,\mathrm{d}\mu\,\mathrm{d}\beta\nonumber\\
	\leq &\ c_1\int_{\mu_1}^{\infty}\exp\left(-\frac{s(\mu-\mu_1)^2}{2\sigma_1^2}\right)\,\mathrm{d}\mu \cdot 2c_2\int_{\tau}^\infty \frac{1}{\prob\left(\frac{1}{\tau_{1,t}}-\sigma_1^2\leq \varepsilon\,\big|\,\widehat{\sigma}_{i,s}=\beta\right)}\beta^{\frac{s}{2}-1} e^{-\frac{\beta}{2\sigma_1^2}}\,\mathrm{d}\beta\nonumber \\
	\leq & \ 2c_2\int_{\tau}^\infty \frac{1}{\prob\left(\frac{1}{\tau_{1,t}}-\sigma_1^2\leq \varepsilon\,\big|\,\widehat{\sigma}_{i,s}=\beta\right)}\beta^{\frac{s}{2}-1} e^{-\frac{\beta}{2\sigma_1^2}}\,\mathrm{d}\beta\nonumber\\
	\leq &\ 2c_2\Gamma\left(\frac{s}{2} \right)\int_{\tau}^\infty\exp\left(\frac{\beta}{2\left(\sigma_1^2+\varepsilon\right)}-\frac{\beta}{2\sigma_1^2}\right)\beta^{\frac{s}{2}-1}\left(1+\frac{\beta}{2\left(\sigma_1^2+\varepsilon\right)}\right)^{-\left(\frac{s}{2}-1\right)}\,\mathrm{d}\beta\label{equ: S-10}\\
	\leq &\ \int_{\tau}^\infty\exp \left(-\frac{\beta\varepsilon}{2\left(\sigma_1^2+\varepsilon\right)\sigma_1^2} \right)\left(\frac{\beta}{2\sigma_1^2}\right)^{\frac{s}{2}-1}\left(1+\frac{\beta}{2\left(\sigma_1^2+\varepsilon\right)}\right)^{-\left(\frac{s}{2}-1\right)}\,\mathrm{d}\beta \nonumber\\
	\leq &\ (\sigma_1^2+\varepsilon)\int_{s}^\infty \exp\left(-\frac{y\varepsilon}{2\sigma_1^2}\right)\left(\frac{\left(\sigma_1^2+\varepsilon\right)y}{2\sigma_1^2+\sigma_1^2y}\right)^{\frac{s}{2}-1}\,\mathrm{d}y \nonumber\\
	\leq &\ C_4\exp\left(-\frac{s\varepsilon}{2}\right)\nonumber
\end{align}
where (\ref{equ: S-10}) follows from Lemma~\ref{lem: S-1}. 
\\\\\textit{Part D:} Use the first case in Lemma~\ref{lem: S-4}, we have 
\[\frac{1-\widetilde{G}_{1s}}{\widetilde{G}_{1s}}\geq \frac{4}{\prob\left(\theta_{1,t}-\mu_1\geq -\varepsilon|\widehat{\mu}_{1,s} = \mu\right)\prob\left(\frac{1}{\tau_{1,t}}-\sigma_1^2\leq \varepsilon|\widehat{\sigma}_{i,s}=\beta\right)}.\] 
Then
\begin{align}
	&\ c_1c_2\int_D \frac{1-\widetilde{G}_{1s}}{\widetilde{G}_{1s}} \exp\left(-\frac{s(\mu-\mu_1)^2}{2\sigma_1^2}\right)\beta^{\frac{s}{2}-1} e^{-\frac{\beta}{2\sigma_1^2}}\,\mathrm{d}\mu\,\mathrm{d}\beta\nonumber\\
	&\le  4c_1c_2\int_D \frac{1}{\prob\left(\theta_{1,t}-\mu_1\geq -\varepsilon\,\big|\,\widehat{\mu}_{1,s} = \mu\right)\prob\left(\frac{1}{\tau_{1,t}}-\sigma_1^2\leq \varepsilon  \,\big|\,\widehat{\sigma}_{i,s}=\beta\right)}\exp\left(-\frac{s(\mu-\mu_1)^2}{2\sigma_1^2}\right)\beta^{\frac{s}{2}-1} e^{-\frac{\beta}{2\sigma_1^2}}\,\mathrm{d}\mu\,\mathrm{d}\beta\nonumber\\
	&\le  2c_1\int_{-\infty}^{\mu_1}\frac{1}{\prob\left(\theta_{1,t}-\mu_1\geq -\varepsilon\,\big|\,\widehat{\mu}_{1,s} = \mu\right)}\exp\left(-\frac{s(\mu-\mu_1)^2}{2\sigma_1^2}\right)\mathrm{d}\mu\nonumber\\
	&\qquad \cdot 2c_2\int_{\tau}^\infty \frac{1}{\prob\left(\frac{1}{\tau_{1,t}}-\sigma_1^2\leq \varepsilon\,\big|\,\widehat{\sigma}_{i,s}=\beta\right)}\beta^{\frac{s}{2}-1} e^{-\frac{\beta}{2\sigma_1^2}}\,\mathrm{d}\beta \nonumber\\
	&\le  C_5 \exp\left(-\frac{s\varepsilon^2}{2}-\frac{s\varepsilon}{2}\right)\label{equ: S-11} .
\end{align}
For (\ref{equ: S-11}), we can reuse the integrations in \textit{Part B} and \textit{Part C}.
\\\\Combine these four parts, we obtain an upper bound of (\ref{equ: S-6}) as follows,
\begin{align*}
	&\E\left[\frac{1}{G_{1s}}-1\right]\leq C_1 \exp\left(-\frac{s\varepsilon^2}{4} \right) + C_3 \exp\left(-\frac{s\varepsilon^2}{2} \right)+C_4\exp\left(-\frac{s\varepsilon}{2}\right) + C_5 \exp\left(-\frac{s\varepsilon^2}{2}-\frac{s\varepsilon}{2}\right)
\end{align*}
 Summing over $s$, we have 
\[\sum_{s=1}^\infty\E\left[\frac{1}{G_{1s}}-1\right]\leq \frac{C_6}{\varepsilon^2} + \frac{C_7}{\varepsilon} + C_8\]

\subsection{Bounding the second term of (\ref{equ: equ15})}
\label{sec: S-2-4}
Following from Lemma \ref{lem: S-3}, we have the following inclusions: \[\left\{\widehat{\mu}_{is} +\sqrt{\frac{2\log (2n)}{s}}\leq \mu_1 - \varepsilon\right\}\subseteq\left\{ \exp\left(-s\left(\Gamma_i-\varepsilon\right)^2\right)\leq \frac{1}{2n}\right\}\]and
\[\left\{\frac{\widehat{\sigma}_i^2}{\sigma_1^2+\varepsilon}\geq h_+^{-1}\left(\frac{\log (2n)}{s}\right) \right\}\cup\left\{  \frac{\widehat{\sigma}_i^2}{\sigma_1^2+\varepsilon}\leq h_-^{-1}\left(\frac{\log (2n)}{s}\right)\right\}\subseteq \left\{\exp\left(-sh\left(\frac{\widehat{\sigma}_i^2}{\sigma_1^2+ \varepsilon}\right)\right)\leq \frac{1}{2n}\right\}\]
where $h_+^{-1}(y) = \max \{x: h(x) = y\}$, and  $h_-^{-1}(y) = \min \{x: h(x) = y\}$.

Hence for $$s\geq u = \max\left\{\frac{2\log (2n)}{\left(\Gamma_i-2\varepsilon\right)^2},\frac{\log (2n)}{h(\sigma_i^2/\sigma_1^2)}\right\},$$ we have 
\begin{align*}
	\prob\left(G_{is}>\frac{1}{n}\right) &\leq \prob\left(\widehat{\mu}_{is} +\sqrt{\frac{2\log (2n)}{s}}\geq \mu_1 - \varepsilon\right) + \prob\left(h_-^{-1}\left(\frac{\log (2n)}{s}\right)\leq \frac{\widehat{\sigma}_i^2}{\sigma_1^2+\varepsilon}\leq h_+^{-1}\left(\frac{\log (2n)}{s}\right)\right) \\
	& \leq \prob\left(\widehat{\mu}_{i,s} - \mu_i\geq  \Gamma_i-\varepsilon - \sqrt{\frac{2\log (2n)}{s}}\right)  + \prob\left(\widehat{\sigma}_i^2 \leq \left(\sigma_1^2+\varepsilon\right)h_+^{-1}\left(\frac{\log (2n)}{s}\right)\right)\\
	& \leq \exp\left(-\frac{s\left(\Gamma_i-\varepsilon - \sqrt{\frac{2\log (2n)}{s}}\right)^2}{2\sigma_i^2}\right) + \exp\left(-(s-1)\frac{\left(\left(\sigma_1^2+\varepsilon\right)h_+^{-1}\left(\frac{\log (2n)}{s}\right)-\sigma_i^2\right)^2}{4\sigma_i^4}\right) \\
	& \leq \exp\left(-\frac{s\varepsilon^2}{\sigma_i^2}\right) + \exp\left(-(s-1)\frac{\varepsilon^2}{\sigma_1^4}\right).
\end{align*}
Summing over $s$,
\begin{align*}
	\sum_{s=1}^n \prob\left(G_{is}\geq 1/n\right) &\leq u + \sum_{s=\lceil u\rceil}^n \exp\left(-\frac{s\varepsilon^2}{\sigma_i^2}\right) + \exp\left(-(s-1)\frac{\varepsilon^2}{\sigma_1^4}\right) \\
	& \leq 1 + \max\left\{\frac{2\log (2n)}{\left(\Gamma_i-2\varepsilon\right)^2},\frac{\log (2n)}{h(\frac{\sigma_i^2}{\sigma_1^2})}\right\} +  \frac{2}{\varepsilon^2}.\end{align*}

Combining the two previous bounds, we have the following lemma,
\begin{lemma} 
\label{lem: S-5}
We have 
	\[\E[T_{i,n}]\leq  1 + \max\left\{\frac{2\log (2n)}{\left(\Gamma_i-2\varepsilon\right)^2},\frac{\log (2n)}{h(\frac{\sigma_i^2}{\sigma_1^2})}\right\} + \frac{C_6}{\varepsilon^2} + \frac{C_7}{\varepsilon} + C_8\]
\end{lemma}
{The finite-time regret bound for MVTS follows from Lemma \ref{lem: S-5} and equation~(\ref{equ: equ10}) in main text.
\begin{theorem} 
\label{thm: S-1}
The finite-time expected regret of MVTS for mean-variance Gaussian bandits satisfies
	\begin{align}
	\E \big[\mathcal{\widetilde{\mathcal{R}}}_n\left(\mathrm{MVTS}\right)\big] 
		&\leq \sum_{i=2}^K  \Bigg(1 + \max\bigg\{\frac{2\log (2n)}{\left(\Gamma_i-2(\log n)^{-1/4}\right)^2},\frac{\log (2n)}{h(\frac{\sigma_i^2}{\sigma_1^2})}\bigg\} \nonumber\\
		&\qquad+ C_6\left(\log n\right)^{1/2} + C_7\left(\log n\right)^{1/4} + C_8 \Bigg)\left(\Delta_i+2\Gamma_{i, \max}^2\right). \nonumber	
		\end{align}
\end{theorem}
Let $\varepsilon = (\log n)^{-\frac{1}{4}}$ and $n\rightarrow +\infty$, the regret bound in {Theorem~\ref{thm: thm3}} follows from Theorem \ref{thm: S-1}.}
\section{Proof of Theorem~\ref{thm: thm1}}
\label{sec: S-3}
The proof is similar to that for proof of Theorem~\ref{thm: thm3}.  For Theorem~\ref{thm: thm1} (MTS), we define following event and conditional probability,\begin{equation}E_i\left(t\right)=\left\{\widehat{\MV}_{i,t} = \rho\theta_{i,t}-\widehat{\sigma}_{i,T_{i,t}}^2\leq \MV_1-(1+\rho)\varepsilon\right\}, \quad G_{is} = \prob\left(E_i(t)^c|T_{i,t}=s\right) \nonumber\end{equation}
\begin{lemma} 
\label{lem: S-6}
we have
\[\prob\left(\widehat{\MV}_{i,t} \geq \normalfont{\MV}_1-(1+\rho)\varepsilon\,\Big|\,\hat{\mu}_i = \mu,T_{i,t}=s\right) \leq \exp\left(-\frac{s\left(\normalfont{\MV}_1-\rho\mu-(1+\rho)\varepsilon\right)^2}{2\rho^2}\right)\]
\end{lemma}
\textit{Proof:} Consider,
\begin{align*}
	&\prob\left(\widehat{\MV}_{i,t} \geq \normalfont{\MV}_1-(1+\rho)\varepsilon\,\Big|\,\hat{\mu}_i = \mu,T_{i,t}=s\right) \\
	&= \prob\left(\rho\theta_{i,t}-\widehat{\sigma}_{i,s}^2\geq \rho\mu_1-\sigma_1^2-(1+\rho)\varepsilon\,\big|\, \hat{\mu}_i = \mu,T_{i,t}=s\right) \\
	&\le \prob\left(\rho\theta_{i,t}\geq \rho\mu_1-\sigma_1^2-(1+\rho)\varepsilon\,\big|\,\hat{\mu}_i = \mu,T_{i,t}=s\right) \\
	&\le  \exp\left(-\frac{s\left(\normalfont{\MV}_1-\rho\mu-(1+\rho)\varepsilon\right)^2}{2\rho^2}\right)
\end{align*}
This lemma is used to bound the second term of (\ref{equ: equ15}) in Lemma~\ref{lemma: lem2}. We also need a lower bound of $G_{1s}$ to bound the first term of (\ref{equ: equ15}) in Lemma~\ref{lemma: lem2}. 
\begin{lemma}
We have
\[\prob\left(\widehat{\MV}_{1,t} \geq \normalfont{\MV}_1-(1+\rho)\varepsilon\,\Big|\,\hat{\mu}_1 = \mu,T_{1,t}=s\right)\geq \begin{cases}
 \frac{1}{2}\prob(\theta_{1,t}\geq \mu_1-\varepsilon|\hat{\mu}_i = \mu,T_{1,t}=s)&\mathrm{ if }\	\widehat{\mu}_{i,s}\leq\mu_1\\
 \frac{1}{4}&\mathrm{ if }\	\widehat{\mu}_{i,s}>\mu_1 
 \end{cases}.
\]
\label{lem: S-7}
\end{lemma}
\textit{Proof:} By direct calculation,
\begin{align}
	&\prob\left(\widehat{\MV}_{1,t} \geq \normalfont{\MV}_1-(1+\rho)\varepsilon\,\Big|\,\hat{\mu}_1 = \mu,T_{1,t}=s\right)\nonumber \\
	&= \prob\left(\rho\theta_{1,t}-\widehat{\sigma}_{1,s}^2\geq \rho\mu_1-\sigma_1^2-(1+\rho)\varepsilon\,\big|\, \hat{\mu}_i = \mu,T_{1,t}=s\right)\nonumber \\
	&\ge \prob\left(\theta_{1,t}\geq \mu_1-\varepsilon\,\big|\,\hat{\mu}_1 = \mu,T_{1,t}=s \right)\prob\left(\widehat{\sigma}_{1,s}^2\leq \sigma_1^2+\varepsilon \right)\nonumber\\
	&\ge \frac{1}{2}\prob \left(\theta_{1,t}\geq \mu_1-\varepsilon\,\big|\,\hat{\mu}_i = \mu,T_{1,t}=s \right) \label{equ: S-12}
\end{align}
Then Lemma~\ref{lem: S-7} is proved by the inequality in (\ref{equ: S-12}) and the following fact: For $X$ being an Gaussian random variable with mean $\mu$ and variance $\sigma^2$, if $x'<\mu $
\[
\Pr(X>x')\geq \frac{1}{2}
\]
\subsection{Bounding the first term of (\ref{equ: equ15})}
\label{sec: S-3-1}
With Lemma~\ref{lem: S-6}, Lemma~\ref{lem: S-7}, we  can now prove   Theorem~\ref{thm: thm1}.

Let $c = 1/\sqrt{2\pi\sigma_1^2}$ and fix $\varepsilon>0$. We will condition on $\hat{\mu}_{i,s}$ and use the same proof technique as that for Theorem~\ref{thm: thm3}. The parameter space $(-\infty,\infty)$ will be divided into two parts, $(-\infty, \infty) = A\cup B$ where 
$$
A = (-\infty, \mu_1-\varepsilon), \quad\mbox{and}\quad B = [\mu_1-\varepsilon, \infty).
$$
We define the conditional version of $G_{1s}$ as  \[\widetilde{G}_{1s} = G_{1s} |_{\widehat{\mu}_{1,s} = \mu, \widehat{\sigma}_{1,s}^2 = \beta}= \prob\left(\widehat{\MV}_{i,t}\geq \MV_1-(1+\rho)\varepsilon\,\big|\,\widehat{\mu}_{1,s} = \mu\right)\]
Consider,
\begin{align*}
	\E\left[\frac{1}{G_{1s}}-1\right] & = c\int_{-\infty}^\infty \frac{1-\widetilde{G}_{1s}}{\widetilde{G}_{1s}} \exp\left(-\frac{s(\mu-\mu_1)^2}{2\sigma_1^2}\right)\, \mathrm{d}\mu\\
	& \leq 4c\int_A \left(1-\widetilde{G}_{1s}\right) \exp\left(-\frac{s(\mu-\mu_1)^2}{2\sigma_1^2}\right)\, \mathrm{d}\mu + 2c\int_B \frac{\exp\left(-\frac{s(\mu-\mu_1)^2}{2\sigma_1^2}\right)}{\prob\left(\theta_{1,t}\geq \mu_1-\varepsilon\,\big|\,\hat{\mu}_i = \mu,T_{1,t}=s \right)}\, \mathrm{d}\mu\\
	&\leq C_9\exp\left(-s \varepsilon^{4} / 2\right)
\end{align*}
We have computed the same integration in (\ref{equ: S-9}) and (\ref{equ: S-8}). Summing over $s$, we have 
\[\sum_{s=1}^\infty\E\left[\frac{1}{G_{1s}}-1\right]\leq \frac{4C_9}{\varepsilon^2}.\]


\subsection{Bounding the second term of (\ref{equ: equ15})}
\label{sec: S-3-2}
Following Lemma \ref{lem: S-6}, \[\left\{\hat{\mu}_{is} +\sqrt{\frac{2\log n}{s}}\leq \frac{\MV_1-(1+\rho)\varepsilon}{\rho}\right\}\subseteq\left\{ G_{is}\leq \frac{1}{n}\right\}\]
Hence for $s\geq u = \frac{2\rho^2\log n}{\left(\rho \Gamma_i -\sigma_1^2-(1+\rho)\varepsilon\right)^2}$, we have 
\begin{align*}
	\prob\left(G_{is}>\frac{1}{n}\right) &\leq \prob\left(\hat{\mu}_{is} +\sqrt{\frac{2\log n}{s}}\geq \frac{\MV_1-(1+\rho)\varepsilon}{\rho}\right) \\
	& = \prob\left(\hat{\mu}_{is} - \mu_i\geq  \Gamma_i-\frac{\sigma_1^2+(1+\rho)\varepsilon}{\rho} - \sqrt{\frac{2\log n}{s}}\right) \\
	& \leq \exp\left(-\frac{s\left(\Gamma_i-\frac{\sigma_1^2+(1+\rho)\varepsilon}{\rho} - \sqrt{\frac{2\log n}{s}}\right)^2}{2\sigma_i^2}\right)
\end{align*}
Summing over $s$,
\begin{align*}
	\sum_{s=1}^n \prob\left(G_{is}\geq 1/n\right) &\leq u + \sum_{s=\lceil u\rceil}^n \exp\left(-\frac{s\left(\Gamma_i-\frac{\sigma_1^2+(1+\rho)\varepsilon}{\rho} - \sqrt{\frac{2\log n}{s}}\right)^2}{2\sigma_i^2}\right) \\
	& \leq 1 + \frac{2\rho^2\log n}{\left(\rho \Gamma_i -\sigma_1^2-(1+\rho)\varepsilon\right)^2} + \frac{2\sigma_i^2}{\left(\Gamma_i - \frac{\sigma_1^2+(1+\rho)\varepsilon}{\rho}\right)^2}\left(\sqrt{\pi\sigma_1^2\log n} + 1\right)
\end{align*}

Combining the two previous bounds, we have the following lemma.
\begin{lemma} 
\label{lem: S-8}
If $\rho > \max\left\{\frac{\sigma_1^2}{\Gamma_i},i=1,2,\cdots,K\right\}$, we have 
	\[\E[T_i\left(n\right)]\leq  \frac{2\rho^2\log n}{\left(\rho \Gamma_i -\sigma_1^2-(1+\rho)\varepsilon\right)^2} + \frac{2}{\sigma_i^2\left(\Gamma_i - \frac{\sigma_1^2+(1+\rho)\varepsilon}{\rho}\right)^2}\left(\sqrt{\pi\sigma_1^2\log n} + 1\right) +  \frac{4C_9}{\varepsilon^2} + 2\]
\end{lemma}
{The finite-time regret bound follows from Lemma \ref{lem: S-8} and equation~(\ref{equ: equ10}) in main text.
\begin{theorem} 
\label{thm: S-2}
The finite-time expected regret of MVTS for mean-variance Gaussian bandits satisfies
	\begin{equation}
	\E \big[\mathcal{\widetilde{\mathcal{R}}}_n\left(\mathrm{MVTS}\right)\big] 
		\leq \sum_{i=2}^K \left(\frac{2\rho^2\log n}{\left(\rho \Gamma_i -\sigma_1^2-(1+\rho)(\log n)^{-1/4}\right)^2} +  4C_9(\log n)^{1/2} + 2 \right)\left(\Delta_i+2\Gamma_{i, \max}^2\right). 
		\nonumber\end{equation}
\end{theorem}
Let $\varepsilon = (\log n)^{-\frac{1}{4}}$ and $n\rightarrow +\infty$, the regret bound in {Theorem~\ref{thm: thm1}} follows from Theorem \ref{thm: S-2}.}

\section{Proof of Theorem~\ref{thm: thm2}}
\label{sec: S-4}
This is also similar to the proof of Theorem~\ref{thm: thm3}.  For Theorem~\ref{thm: thm2} (VTS), we define following event and conditional probability,\begin{equation}E_i\left(t\right)=\left\{\widehat{\MV}_{i,t} = \rho\widehat{\mu}_{i,s}-\frac{1}{\tau_{i,t}}\leq \MV_1-(1+\rho)\varepsilon\right\}, \quad G_{is} = \prob\left(E_i(t)^c|T_{i,t}=s\right). \nonumber\end{equation}

\begin{lemma} 
\label{lem: S-9}
Given $\hat{\sigma}_{i,s}^2 = \sigma^2$ and $ T_{i,t} = s $ such that $$s > \frac{2\sigma_i^2\log (2n)}{(\Gamma_{1,i}-\varepsilon)^2},$$ we have
\[\prob\left(\widehat{\MV}_{i,t} \geq \normalfont{\MV}_1-(1+\rho)\varepsilon\,\Big|\,\hat{\sigma}_{i,s}^2 = \sigma_1^2,T_{i,t}=s\right) \leq \frac{1}{2n} + \exp\left(-sh \left(\frac{\sigma^2}{\sigma_1^2+\varepsilon} \right)\right) \]
\end{lemma}
\textit{Proof:} Consider,
\begin{align*}
	&\prob\left(\widehat{\MV}_{i,t} \geq \normalfont{\MV}_1-(1+\rho)\varepsilon\,\Big|\,\hat{\sigma}_{i,s}^2 = \sigma^2,T_{i,t}=s\right) \\
	&=\prob\left(\rho\widehat{\mu}_{i,s}-\frac{1}{\tau_{i,t}}\geq \rho\mu_1-\sigma_1^2-(1+\rho)\varepsilon\,\Big|\, \hat{\sigma}_{i,s}^2 = \sigma^2,T_{i,t}=s\right) \\
	&\le \prob\left(\rho\widehat{\mu}_{i,s}\geq \rho\mu_1-\rho\varepsilon\right) + \prob\left(\frac{1}{\tau_{i,t}}\leq \sigma_1^2+\varepsilon\,\Big|\,\hat{\sigma}_{i,s}^2 = \sigma^2, T_{i,t} =s\right) \\
	&\le \frac{1}{2n} + \exp\left(-sh \left(\frac{\sigma^2}{\sigma_1^2+\varepsilon} \right)\right) .
	\end{align*}
This lemma is used to bound the second term of (\ref{equ: equ15}) in Lemma~\ref{lemma: lem2}.
 We need also a lower bound of $G_{1s}$ to bound the first term of (\ref{equ: equ15}) in Lemma~\ref{lemma: lem2}. 
\begin{lemma}
\label{lem: S-10}
Given $\hat{\mu}_{1,s} = \mu$ and $ T_{1,t} = s$, we have
\[\prob\left(\widehat{\MV}_{1,t} \geq \normalfont{\MV}_1-(1+\rho)\varepsilon\,\big|\,\hat{\sigma}_{i,s}^2 = \sigma^2, T_{i,t} =s\right)\geq \begin{cases}
 \frac{1}{2}\prob(\frac{1}{\tau_{1,t}}\leq \sigma_1^2+\varepsilon\, \big|\,\hat{\sigma}_{i,s}^2 = \sigma^2, T_{1,t} =s) &\mathrm{ if }\	\hat{\sigma}_{i,s}^2\leq\sigma_1^2\\
 \frac{1}{4}&\mathrm{ if }\	\hat{\sigma}_{i,s}^2>\sigma_1^2 
 \end{cases}  .
\]
\end{lemma}
\textit{Proof:} By direct calculation,
\begin{align}
	&\prob\left(\widehat{\MV}_{1,t} \geq \normalfont{\MV}_1-(1+\rho)\varepsilon \,\Big|\,\hat{\mu}_1 = \mu,T_{1,t}=s\right)\nonumber \\
	&= \prob\left(\rho\widehat{\mu}_{1,s}-\frac{1}{\tau_{1,t}}\geq \rho\mu_1-\sigma_1^2-(1+\rho)\varepsilon\,\Big|\, \hat{\sigma}_{1,s}^2 = \sigma^2,T_{1,t}=s\right)\nonumber \\
	&\ge \prob\left(\frac{1}{\tau_{1,t}}\leq \sigma_1^2+\varepsilon\,\Big|\,\hat{\sigma}_{i,s}^2 = \sigma^2, T_{1,t} =s\right)\prob\left(\widehat{\mu}_{1,s}\geq \mu_1-\varepsilon\right)\nonumber \\
	&\ge\frac{1}{2}\prob\left(\frac{1}{\tau_{1,t}}\leq \sigma_1^2+\varepsilon\,\Big|\,\hat{\sigma}_{i,s}^2 = \sigma^2, T_{1,t} =s \right).\label{equ: S-13}
\end{align}
Then Lemma~\ref{lem: S-10} is proved by the inequality in (\ref{equ: S-13}) and the following fact: If $X$ is an inverse-Gamma random variable with shape $\alpha$ and rate $\beta$, if $x>\frac{\beta}{\alpha-1} $
\[
\Pr(X<x)\geq \frac{1}{2}.
\]

\subsection{Bounding the first term of (\ref{equ: equ15})}

\label{sec: S-4-1}

Let $c = \frac{1}{2^{s/2} \Gamma\left(s/2\right)\sigma_1^{s-2}}$ and $\tau = s(\sigma_1^2+\varepsilon)$ for some fixed $\varepsilon>0$. To calculate the expectation  conditioned on $\widehat{\sigma}_{i,s}^2$,  we will use the same proof technique as the proof of Theorem~\ref{thm: thm3}. In particular,  the parameter space $(0,\infty)$ will be divided into two parts, i.e., $ (0, \infty) = A\cup B$ where 
\[A = (0, \tau),\quad\mbox{and}\quad B = [\tau, \infty).\] 
We define the conditional version of $G_{1s}$ as  \[\widetilde{G}_{1s} = G_{1s} |_{\widehat{\mu}_{1,s} = \mu, \widehat{\sigma}_{1,s}^2 = \beta}= \prob\left(\widehat{\MV}_{i,t}\geq \MV_1-(1+\rho)\varepsilon\,\big|\, \widehat{\sigma}_{1,s}^2 = \beta\right)\]
Then
\begin{align*}
	\E\left[\frac{1}{G_{1s}}-1\right] & = c\int_{0}^\infty \frac{1-\widetilde{G}_{1s}}{\widetilde{G}_{1s}} \beta^{\frac{s}{2}-1} e^{-\frac{\beta}{2\sigma_1^2}}\mathrm{d}\beta\\
	& \leq 4c\int_A (1-\widetilde{G}_{1s})\beta^{\frac{s}{2}-1} e^{-\frac{\beta}{2\sigma_1^2}}\mathrm{d}\beta + 2c\int_B \frac{1}{\prob(\frac{1}{\tau_{1,t}}\leq \sigma_1^2+\varepsilon|\hat{\sigma}_{i,s}^2 = \sigma^2, T_{1,t} =s)}\beta^{\frac{s}{2}-1} e^{-\frac{\beta}{2\sigma_1^2}}\mathrm{d}\beta \\
	& \leq 4c\int_A\exp\left(-\frac{\left(\beta-s\left(\sigma_1^2+\varepsilon\right)\right)^2}{4s\left(\sigma_1^2+\varepsilon\right)^2}\right)\beta^{\frac{s}{2}-1} e^{-\frac{\beta}{2\sigma_1^2}}\mathrm{d}\beta \\
	&\qquad + 2c\Gamma(\frac{s}{2})\int_B\exp\left(\frac{\beta}{2\left(\sigma_1^2+\varepsilon\right)}-\frac{\beta}{2\sigma_1^2}\right)\beta^{\frac{s}{2}-1}\left(1+\frac{\beta}{2\left(\sigma_1^2+\varepsilon\right)}\right)^{-\left(\frac{s}{2}-1\right)}\mathrm{d}\beta\\
	& \leq 4\exp\left(-\frac{s\varepsilon^2}{4\sigma_1^2}\right) + \int_B\exp\left(-\frac{\beta\varepsilon}{2\left(\sigma_1^2+\varepsilon\right)\sigma_1^2}\right)\left(\frac{\beta}{2\sigma_1^2}\right)^{\frac{s}{2}-1}\left(1+\frac{\beta}{2\left(\sigma_1^2+\varepsilon\right)}\right)^{-\left(\frac{s}{2}-1\right)}\mathrm{d}\beta \\
	& \leq 4\exp\left(-\frac{s\varepsilon^2}{4\sigma_1^2}\right) + (\sigma_1^2+\varepsilon)\int_{s}^\infty \exp\left(-\frac{y\varepsilon}{2\sigma_1^2}\right)\left(\frac{\left(\sigma_1^2+\varepsilon\right)y}{2\sigma_1^2+\sigma_1^2y}\right)^{\frac{s}{2}-1}\mathrm{d}\beta \\
	& \leq C_{10}\exp\left(-\frac{s\varepsilon^2}{4}\right) + C_{11}\exp\left(-\frac{s\varepsilon}{2}\right).
\end{align*}
We omit the details because the same integrations have been computed in (\ref{equ: S-10}) and (\ref{equ: S-11}). Summing from $s=0$ to $\infty$ shows that 
\[\sum_{s=0}^\infty\E\left[\frac{1}{G_{1s}}-1\right] \leq \frac{4C_{10}}{\varepsilon^2} + \frac{2C_{11}}{\varepsilon}.\]

\subsection{Bounding the second term of (\ref{equ: equ15})}
\label{sec: S-4-2}
Hence, similar to the analysis of MVTS,
\[\left\{\frac{\widehat{\sigma}_i^2}{\sigma_1^2+\varepsilon}\geq h_+^{-1}\left(\frac{\log (2n)}{s}\right) \right\}\cup\left\{  \frac{\widehat{\sigma}_i^2}{\sigma_1^2+\varepsilon}\leq h_-^{-1}\left(\frac{\log (2n)}{s}\right)\right\}\subseteq\left\{ \exp\left(-sh\left(\frac{\widehat{\sigma}_i^2}{\sigma_1^2+ \varepsilon}\right)\right)\leq \frac{1}{2n}\right\}.\]
Then if $s>\frac{\log (2n)}{h(\sigma_i^2/\sigma_1^2)} = s^*$, $\rho\leq \frac{\Delta_i}{\Gamma_i}$, and $ \Gamma_i^2\geq 8\sigma_1^2h\left(\frac{\sigma_i^2}{\sigma_1^2}\right)$
\begin{align*}
	\prob\left(G_{is}>\frac{1}{n}\right) &\leq \prob\left(h_-^{-1}\left(\frac{\log (2n)}{s}\right)\leq \frac{\widehat{\sigma}_i^2}{\sigma_1^2+\varepsilon}\leq h_+^{-1}\left(\frac{\log (2n)}{s}\right)\right) \\
	& \leq \prob\left(\widehat{\sigma}_i^2 \leq \left(\sigma_1^2+\varepsilon\right)h_+^{-1}\left(\frac{\log (2n)}{s}\right)\right)\\
	& \leq  \exp\left(-(s-1)\frac{\left(\left(\sigma_1^2+\varepsilon\right)h_+^{-1}\left(\frac{\log (2n)}{s}\right)-\sigma_i^2\right)^2}{4\sigma_i^4}\right) \\
	& \leq \exp\left(-(s-1)\frac{\varepsilon^2}{\sigma_1^4} \right).
\end{align*}
Hence \[\sum_{s=1}^n \prob\left(G_{is}>\frac{1}{n} \right)\leq s^*+2\sum_{s = \left\lceil s^*\right\rceil}^n\exp\left(-\left(s-1\right)\frac{\varepsilon^2}{\sigma_1^4}\right)= \frac{\log (2n)}{ h\left(\frac{\sigma_i^2}{\sigma_1^2}\right)} +1 + \frac{1}{\varepsilon^2}.\]
\begin{lemma}
\label{lem: S-11}
The number of times that VTS pulls arm $i$ is bounded as
	\[\E[T_{i,n}]\leq \frac{\log (2n)}{ h\left(\frac{\sigma_i^2}{\sigma_1^2}\right)}  + \frac{4C_{10}}{\varepsilon^2} + \frac{2C_{11}}{\varepsilon} + C_{12}\]
\end{lemma}
{
The finite-time regret bound follows from Lemma \ref{lem: S-11} and equation~(\ref{equ: equ10}) in main text, 
\begin{theorem} 
\label{thm: S-3}
The finite-time expected regret of MVTS for mean-variance Gaussian bandits satisfies
	\begin{equation}
	\E \big[\mathcal{\widetilde{\mathcal{R}}}_n\left(\mathrm{MVTS}\right)\big] 
		\;\;\leq \sum_{i=2}^K \left(\frac{\log (2n)}{ h\left(\frac{\sigma_i^2}{\sigma_1^2}\right)}  + 4C_{10} (\log n)^{1/2} + 2C_{11} (\log n)^{1/4} + C_{12} \right)\left(\Delta_i+2\Gamma_{i, \max}^2\right). \nonumber\end{equation}
\end{theorem}
Let $\varepsilon = (\log n)^{-\frac{1}{4}}$ and $n\rightarrow +\infty$, the regret bound in {Theorem~\ref{thm: thm2}} follows from Theorem \ref{thm: S-3}.}

%
\section{Proof of Theorem \ref{thm: thm4}}
We provide following useful lemmas before we process to prove the theorem.
\begin{lemma}[Chernoff-Hoeffding bound I]
\label{lem: ch1}
	Let $X_1,\ldots,X_n$ be independent $\{0,1\}$-valued random variables (i.e., Bernoulli random variables)  with  $\E[X_i]=p_i$. Let $X = \frac{1}{n}\sum_{i=1}^nX_i, \mu = \E[X]=\frac{1}{n}\sum_{i=1}^n p_i$. Then, for any $0<\lambda < 1-\mu$,
	\[\mathbb{P}(X \geq \mu+\lambda) \leq \exp (-n d(\mu+\lambda, \mu))\]
	and, for any $0<\lambda<\mu$,
	\[\mathbb{P}(X \leq \mu-\lambda) \leq \exp (-n d(\mu-\lambda, \mu))\]
	where $d(a,b)=a\log \frac{a}{b}+(1-a)\log \frac{1-a}{1-b}$.
\end{lemma}
\begin{lemma}[Chernoff-Hoeffding bound II]
\label{lem: ch2}
	Let $X_1,\cdots,X_n$ be random variables with common range $[0,1]$ and such that $\E[X_t|X_1,\ldots,X_{t-1}]=\mu$. Let $S_n = X_1+\ldots+X_n$. Then for all $a\geq 0$,
	\[\mathbb{P}\left(S_{n} \geq n \mu+a\right) \leq e^{-2 a^{2} / n},\]
	\[\mathbb{P}\left(S_{n} \leq n \mu-a\right) \leq e^{-2 a^{2} / n}.\]
\end{lemma}
\begin{lemma}[Relationship between Beta distribution and Binomial distribution]
	For all positive integers $\alpha,\beta$,
	\[F_{\alpha, \beta}^{\mathrm{Beta}}(y)=1-F_{\alpha+\beta-1, y}^{\mathrm{B}}(\alpha-1).\]
\end{lemma}
\begin{lemma}
\label{lem: para_concentration}
If $M$ is a  Binomial random variable with $s$ trials and probability of success $p$, then
	\begin{equation}
     \mathbb{P}\left( d(\tilde{p} -\varepsilon ,M/s )\le\frac{\log(2n)}{s} \right) \leq \exp\left( -2s \Big( \tilde{p} - p -\varepsilon \pm\sqrt{ \frac{ \log(2n)}{s}} \Big)^2 \right).\nonumber
\end{equation}
where the $\pm$ is taken so that the exponent is minimized.
\end{lemma}
\textit{Proof}:
Let us define
\begin{equation}
\mathrm{p}_s:=\mathbb{P}\left( d(\tilde{p} -\sqrt{\varepsilon} ,M/s )\le\frac{\log(2n)}{s} \right).\nonumber
\end{equation}
Clearly by the law of large numbers $\mathrm{p}_s\to 0$ as $s\to\infty$. 
We can write 
\begin{equation}
\mathrm{p}_s:=\mathbb{P}\left(  d \Big(\tilde{p} -\sqrt{\varepsilon} ,\frac{1}{s}\sum_{i=1}^s X_i \Big)\le\frac{\log(2n)}{s}  \right),\nonumber
\end{equation}
where $X_i$ are i.i.d.\ Bernoulli random variables with probability of success $p$.   

Now by a slightly strengthened form of Sanov's theorem \citep[Problem 2.12(c)]{Csi97}, we have the large deviations bound
\begin{equation}
\mathrm{p}_s\le \exp\left( -s \min_{q \in\mathcal{A}_s } d(q , p_1) \right)
\nonumber
\end{equation}
where 
\begin{equation}
\mathcal{A}_s := \left\{ q\in [0,1]: d (p_1 -\sqrt{\varepsilon} , q)\le\frac{2\log(2n)}{s}   \right\}.
\nonumber
\end{equation}
By Pinsker's inequality $d(p,q)\ge 2(p-q)^2$ (assuming natural logs) so 
\begin{equation}
\mathcal{A}_s\subset\mathcal{A}_s':= \left\{ q\in [0,1]:  (p_1-\sqrt{\varepsilon}-q)^2\le\frac{ \log(2n)}{s}   \right\}.
\nonumber
\end{equation}
Hence, one has
\begin{equation}
\mathrm{p}_s\le \exp\left( -s \min_{q \in\mathcal{A}_s' } d(q , p_1) \right)= \exp\left( -s d\Big( p_1-\sqrt{\varepsilon} \pm\sqrt{ \frac{ \log(2n)}{s}} , p_1\Big) \right).
\nonumber
\end{equation}
The lemma is proven by applying Pinsker's inequality again.

\begin{lemma}[Tail Upper Bound]
\label{lem: S-12}
We have
    \[\prob\left(\widehat{\MV}_{i,s} \geq \MV_1 - \varepsilon\,\Big|\, T_{i,t} = s,\alpha_{i,t}=m \right)\leq \exp\left(-sd\left(x-\sqrt{\varepsilon},\frac{m}{s}\right)\right)  + \exp\left(-sd\left(y+\sqrt{\varepsilon},\frac{m}{s}\right)\right).\]
\end{lemma}
\textit{Proof:}
Let $x = \frac{(1-\rho)+|1-\rho-2p_1|}{2}, y = \frac{(1-\rho)-|1-\rho-2p_1|}{2}$, then
\begin{align*}
	  &\prob\left(\widehat{\MV}_{i,s} \geq \MV_1 - \varepsilon\,\Big|\, T_{i,t} = s,\alpha_{i,t}=m \right)\\
	=\; & \prob\left(\rho\theta_{i,t}-\theta_{i,t}(1-\theta_{i,t})\geq \rho p_1 -p_1(1-p_1)-\epsilon\,\big|\, T_{i,t} = s,\alpha_{i,t}=m \right) \\
	=\; & \prob \left(\theta_{i,t}^2-p_1^2-(1-p)(\theta_{i,t}-p_1)\geq -\epsilon\,\big|\, T_{i,t} = s, \alpha_{i,t}=m \right) \\
	=\; & \prob\left(\theta_{i,t}\geq \frac{(1-\rho)+\sqrt{(1-\rho)^2+4(p_1^2-p_1(1-\rho)-\varepsilon)}}{2}\right) \\
	& \hspace{2cm} + \prob\left(\theta_{i,t}\leq \frac{(1-\rho)-\sqrt{(1-\rho)^2+4(p_1^2-p_1(1-\rho)-\varepsilon)}}{2}\right)\\
	\leq\; &  \prob\left(\theta_{i,t}\geq \frac{(1-\rho)+|1-\rho-2p_1|}{2}-\sqrt{\varepsilon}\right)  + \prob\left(\theta_{i,t}\leq \frac{(1-\rho)-|1-\rho-2p_1|}{2}+\sqrt{\varepsilon}\right) \\
	\leq\; & \exp\left(-sd\left(x-\sqrt{\varepsilon},\frac{m}{s}\right)\right) + \exp\left(-sd\left(y+\sqrt{\varepsilon},\frac{m}{s}\right)\right).
\end{align*}
The last inequality follows from Chernoff-Hoeffding bound (Lemma \ref{lem: ch1}).
\begin{lemma}[Tail Lower Bound]
\label{lem: S-13}
We have
    \[\prob\left(\widehat{\MV}_{i,s} \geq \MV_1 - \varepsilon\,\Big|\, T_{i,t} = s, \alpha_{i,t}=m \right)\geq \begin{cases}
	F_{s+1,\tilde{y}}^{\mathrm{B}}(m) & \mathrm{ if }\; 1-\rho-2p_1<0\\
	1-F_{s+1,\tilde{y}}^{\mathrm{B}}(m) & \mathrm{ if }\; 1-\rho-2p_1>0
	\end{cases}.\]
\end{lemma}
where 
\begin{equation}
\tilde{x} = \frac{(1-\rho)+\sqrt{(1-\rho)^2+4(p_1^2-p_1(1-\rho)-\varepsilon)}}{2},\qquad \tilde{y} = \frac{(1-\rho)-\sqrt{(1-\rho)^2+4(p_1^2-p_1(1-\rho)-\varepsilon)}}{2} \nonumber
\end{equation}
and $F_{n,p}^{\mathrm{B}}(\cdot)$ is the cumulative distribution function of the Binomial distribution.
\\\textit{Proof:} Consider,
\begin{align*}
	  &\prob\left(\widehat{\MV}_{i,s} \geq \MV_1 - \varepsilon\,\Big|\, T_{i,t} = s, \alpha_{i,t}=m \right)\\
	=\; & \prob(\rho\theta_{i,t}-\theta_{i,t}(1-\theta_{i,t})\geq \rho p_1 -p_1(1-p_1)-\epsilon\Big|\, T_{i,t} = s,\alpha_{i,t}=m) \\
	=\; & \prob(\theta_{i,t}^2-p_1^2-(1-p)(\theta_{i,t}-p_1)\geq -\epsilon\,\Big|\, T_{i,t} = s,\alpha_{i,t}=m) \\
	=\; & \prob\left(\theta_{i,t}\geq \frac{(1-\rho)+\sqrt{(1-\rho)^2+4(p_1^2-p_1(1-\rho)-\varepsilon)}}{2}\,\Big|\, T_{i,t} = s,\alpha_{i,t}=m\right) \\
	& \hspace{2cm} + \prob\left(\theta_{i,t}\leq \frac{(1-\rho)-\sqrt{(1-\rho)^2+4(p_1^2-p_1(1-\rho)-\varepsilon)}}{2}\,\Big|\, T_{i,t} = s, \alpha_{i,t}=m\right)\\
	=\; & F^{\mathrm{B}}_{s+1,\tilde{x}}(m) + 1-F_{s+1,\tilde{y}}^{\mathrm{B}}(m).
\end{align*}
Then we have following lower bound,
\begin{equation*}
	\prob\left(\widehat{\MV}_{i,s} \geq \MV_1 - \varepsilon\,\Big|\, T_{i,t} = s,S_{i,t}=m \right) \geq 
	\begin{cases}
	F_{s+1,\tilde{x}}^{\mathrm{B}}(m) & \mathrm{ if }\; 1-\rho-2p_1<0\\
	1-F_{s+1,\tilde{y}}^{\mathrm{B}}(m) & \mathrm{ if }\; 1-\rho-2p_1>0
	\end{cases} .
\end{equation*}

\subsection{Bounding the first term of (\ref{equ: equ15})}
With Lemma~\ref{lem: S-12}, Lemma~\ref{lem: S-13}, we can now prove   Theorem~\ref{thm: thm4}.

Fix $\varepsilon>0$. We will condition on $S_{i,s}$ and use the same proof technique as the proof of Theorem~\ref{thm: thm3}.

Consider,
\[\E\left[\frac{1}{G_{1s}}-1\right]  \leq \sum_{m=0}^s \frac{1}{\prob\left(\widehat{\MV}_{i,s} \geq \MV_1 - \varepsilon\,\Big|\, T_{i,t} = s,S_{i,t}=m \right)} {s \choose m}p_1^m(1-p_1)^{s-m}.\]
\textit{Case 1}: If $1-\rho-2p_1>0$, 
\begin{align*}
	\E\left[\frac{1}{G_{1s}}-1\right] & \leq \sum_{m=0}^s \frac{1}{\prob\left(\widehat{\MV}_{i,s} \geq \MV_1 - \varepsilon\,\Big|\, T_{i,t} = s,S_{i,t}=m \right)} {s \choose m}p_1^m(1-p_1)^{s-m} \\
	& \leq \sum_{m=0}^s\frac{1}{1-F_{s+1,\tilde{y}}^{\mathrm{B}}(m)} {s \choose m}p_1^m(1-p_1)^{s-m}\\
	&\leq \sum_{m=0}^{\lfloor \tilde{y}s\rfloor} 2{s \choose m}p_1^m(1-p_1)^{s-m} + \sum_{m=\lfloor \tilde{y}s\rfloor+1}^{s}\frac{{s \choose m}p_1^m(1-p_1)^{s-m}}{{s+1 \choose m}\tilde{y}^m(1-\tilde{y})^{s+1-m}} \\
	& \leq 2\exp(-\frac{2(\lfloor \tilde{y}s\rfloor-sp_1)^2}{s}) + \frac{1}{1-\tilde{y}}\sum_{m=\lfloor \tilde{y}s\rfloor+1}^{s}\frac{p_1^m(1-p_1)^{s-m}}{\tilde{y}^m(1-\tilde{y})^{s-m}} \\
	& \leq 2\exp(-2s(\tilde{y}-p_1)^2) + \frac{p_1}{1-\tilde{y}}\exp(-sd(\tilde{y},p_1)).
\end{align*}
The first part follows from Lemma \ref{lem: ch2}, the second part is by direct computation.
\\\textit{Case 2}: If $1-\rho-2p_1<0$, 
\begin{align*}
	\E\left[\frac{1}{G_{1s}}-1\right] & \leq \sum_{m=0}^s \frac{1}{\prob\left(\widehat{\MV}_{i,s} \geq \MV_1 - \varepsilon\,\Big|\, T_{i,t} = s,S_{i,t}=m \right)} {s \choose m}p_1^m(1-p_1)^{s-m} \\
	& \leq \sum_{m=0}^s\frac{1}{F_{s+1,\tilde{x}}^{\mathrm{B}}(m)} {s \choose m}p_1^m(1-p_1)^{s-m}\\
	&\leq \sum_{m=0}^{\lfloor \tilde{x}s\rfloor}\frac{{s \choose m}p_1^m(1-p_1)^{s-m}}{{s+1 \choose m}\tilde{x}^m(1-\tilde{x})^{s+1-m}}  + \sum_{m=\lfloor \tilde{x}s\rfloor+1}^{s}2{s \choose m}p_1^m(1-p_1)^{s-m} \\
	& \leq \frac{1}{1-\tilde{x}}\sum_{m=0}^{\lfloor \tilde{x}s\rfloor}\frac{p_1^m(1-p_1)^{s-m}}{\tilde{x}^m(1-\tilde{x})^{s-m}} + 2\exp\Big(-\frac{2(\lfloor \tilde{x}s\rfloor + 1-sp_1)^2}{s}\Big) \\
	& \leq \frac{p_1}{1-\tilde{x}}\exp(-sd(\tilde{x},p_1)) + 2\exp\Big(-\frac{2(\lfloor \tilde{x}s\rfloor + 1-sp_1)^2}{s}\Big).
\end{align*}
The first part follows from Lemma \ref{lem: ch2}, the second part is by direct computation.

 Summing over $s$, we have 
\\\textit{Case 1}:
\begin{equation}\label{equ: S-14}
\sum_{s=1}^\infty\E\left[\frac{1}{G_{1s}}-1\right]\leq \frac{C_{13}}{(\tilde{y}-p_1)^2} + \frac{C_{14}}{d(\tilde{y},p_1)}.\end{equation}
\textit{Case 2}:
\begin{equation}\label{equ: S-15}
\sum_{s=1}^\infty\E\left[\frac{1}{G_{1s}}-1\right]\leq \frac{C_{15}}{(\tilde{x}-p_1)^2} + \frac{C_{16}}{d(\tilde{x},p_1)}.\end{equation}
\subsection{Bounding the second term of (\ref{equ: equ15})}
Follow from Lemma \ref{lem: S-12}, we have the following inclusions: \[\left\{d\left(x-\sqrt{\varepsilon},\frac{m}{s}\right)\geq \frac{\log (2n)}{s}\right\}\subseteq\left\{ \exp\left(-sd\left(x-\sqrt{\varepsilon},\frac{m}{s}\right)\right)\leq \frac{1}{2n}\right\}\]and
\[\left\{d\left(y+\sqrt{\varepsilon},\frac{m}{s}\right)\geq \frac{\log (2n)}{s} \right\}\subseteq \left\{\exp\left(-sd\left(y+\sqrt{\varepsilon},\frac{m}{s}\right)\right)\leq \frac{1}{2n}\right\}\]

Hence for $$s\geq u = \max\left\{\frac{\log (2n)}{2\left(\Gamma_i-\sqrt{\varepsilon}\right)^2},\frac{\log (2n)}{2\left(1-\rho-p_1-p_i-\sqrt{\varepsilon}\right)^2}\right\},$$ we have 
\begin{equation}
\label{equ: S-16}
	\prob\left(G_{is}>\frac{1}{n}\right) \leq \prob\left(d\left(x-\sqrt{\varepsilon},\frac{m}{s}\right)\leq \frac{\log (2n)}{s}\right) + \prob\left(d\left(y+\sqrt{\varepsilon},\frac{m}{s}\right)\leq \frac{\log (2n)}{s}\right) .
\end{equation}
\textit{Case 1}: If $1-\rho-2p_1 > 0$, then $x = 1-\rho-p_1>p_1$, $y = p_1$,
then, the first term of (\ref{equ: S-16}) can be bounded by applying Lemma \ref{lem: para_concentration},
\begin{align*}
	&\prob\left(d\left(x-\sqrt{\varepsilon},\frac{m}{s}\right)\leq \frac{\log (2n)}{s}\right)\\
	=\; & \prob\left(d\left(1-\rho-p_1-\sqrt{\varepsilon},\frac{m}{s}\right)\leq \frac{\log (2n)}{s}\right)\\
	\leq\; & \exp\left(-2s\Big(1-\rho-p_1-p_i-\sqrt{\varepsilon} \pm \sqrt{\frac{\log (2n)}{2s}}\Big)^2\right) .
\end{align*}
The second term of (\ref{equ: S-16}) is bounded as follows,
\begin{align*}
	&\prob\left(d\left(y+\sqrt{\varepsilon},\frac{m}{s}\right)\leq \frac{\log (2n)}{s}\right)\\
	=\; & \prob\left(d\left(p_1-\sqrt{\varepsilon},\frac{m}{s}\right)\leq \frac{\log (2n)}{s}\right)\\
	\leq\; & \exp\left(-2s(p_1-p_i-\sqrt{\varepsilon}\pm \sqrt{\frac{\log (2n)}{2s}})^2\right) .
\end{align*}

\textit{Case 2}: If $1-\rho-2p_1 < 0$, then $x = p_1$, $y = 1-\rho-p_1<p_1$,
we then apply Lemma~\ref{lem: para_concentration} again,
\begin{align*}
	&\prob\left(d\left(x-\sqrt{\varepsilon},\frac{m}{s}\right)\leq \frac{\log (2n)}{s}\right)\\
	=\; & \prob\left(d\left(p_1-\sqrt{\varepsilon},\frac{m}{s}\right)\leq \frac{\log (2n)}{s}\right)\\
	\leq\; & \exp\left(-2s\Big(p_1-p_i-\sqrt{\varepsilon}\pm \sqrt{\frac{\log (2n)}{2s}}\Big)^2\right) .
\end{align*}
The second term of (\ref{equ: S-16}) is bounded as follows,
\begin{align*}
	&\prob\left(d\left(y+\sqrt{\varepsilon},\frac{m}{s}\right)\leq \frac{\log (2n)}{s}\right)\\
	=\; & \prob\left(d\left(1-\rho-p_1-\sqrt{\varepsilon},\frac{m}{s}\right)\leq \frac{\log (2n)}{s}\right)\\
	\leq\; & \exp\left(-2s\Big(1-\rho-p_1-p_i-\sqrt{\varepsilon}\pm \sqrt{\frac{\log (2n)}{2s}}\Big)^2\right) .
\end{align*}
Combining  the two cases, we have 
\begin{align*}
	\prob\left(G_{is}>\frac{1}{n}\right) \leq\; & \prob\left(d\left(x-\sqrt{\varepsilon},\frac{m}{s}\right)\leq \frac{\log (2n)}{s}\right) + \prob\left(d\left(y+\sqrt{\varepsilon},\frac{m}{s}\right)\leq \frac{\log (2n)}{s}\right) \\
	\leq\; & \exp\left(-2s \Big(p_1-p_i-\sqrt{\varepsilon}\pm \sqrt{\frac{\log (2n)}{2s}}\Big)^2\right) + \exp\left(-2s\Big(1-\rho-p_1-p_i-\sqrt{\varepsilon}\pm \sqrt{\frac{\log (2n)}{2s}}\Big)^2\right).
\end{align*}
Summing over $s$,
\begin{align}
	\sum_{s=1}^n \prob\left(G_{is}\geq 1/n\right) &\leq u + \sum_{s=\lceil u\rceil}^n \exp\left(-2s \Big(p_1-p_i-\sqrt{\varepsilon}\pm \sqrt{\frac{\log (2n)}{2s}}\Big)^2\right)  \nonumber \\
	&\qquad+ \exp\left(-2s\Big(1-\rho-p_1-p_i-\sqrt{\varepsilon}\pm \sqrt{\frac{\log (2n)}{2s}}\Big)^2\right)  \nonumber\\
	& \label{equ: S-17}\leq \max\left\{\frac{\log (2n)}{2\left(\Gamma_i-\sqrt{\varepsilon}\right)^2},\frac{\log (2n)}{2\left(1-\rho-p_1-p_i-\sqrt{\varepsilon}\right)^2}\right\} +  \frac{C_{17}}{\varepsilon^2} + \frac{C_{18}}{\varepsilon}.\end{align}
\begin{lemma}
\label{lem: S-18}
The expected number of times that BMVTS pulls arm $i$ is bounded as
	\[\E[T_{i,n}]\leq \max\left\{\frac{\log (2n)}{2\left(\Gamma_i-\sqrt{\varepsilon}\right)^2},\frac{\log (2n)}{2\left(1-\rho-p_1-p_i-\sqrt{\varepsilon}\right)^2}\right\} +  \frac{C_{19}}{\varepsilon^2} + \frac{C_{20}}{\varepsilon}.\]
\end{lemma}
\textit{Proof}:
This lemma follows from equations (\ref{equ: S-14}), (\ref{equ: S-15}), (\ref{equ: S-17}), and notice that  as $\varepsilon\to 0^+$, 
\[\frac{1}{(\tilde{x}-p_1)^2} = \Theta \Big(\frac{1}{\varepsilon^2}\Big),\;\; \frac{1}{(\tilde{y}-p_1)^2} = \Theta\Big(\frac{1}{\varepsilon^2}\Big), \;\; \frac{1}{d(\tilde{x},p_1)} = \Theta\Big(\frac{1}{\varepsilon^2}\Big),\;\; \frac{1}{d(\tilde{y},p_1)} = \Theta\Big(\frac{1}{\varepsilon^2}\Big)\]

The finite-time regret bound follows from Lemma \ref{lem: S-18} and equation~(\ref{equ: equ10}) in  the main text, 
\begin{theorem} 
\label{thm: S-4}
The finite-time expected regret of BMVTS for mean-variance Bernoulli bandits satisfies
	\begin{equation}
	\E \big[\mathcal{\widetilde{\mathcal{R}}}_n\left(\mathrm{BMVTS}\right)\big] 
 \leq \sum_{i=2}^K \left(\max\left\{\frac{\log (2n)}{2\left(\Gamma_i-\sqrt{\varepsilon}\right)^2},\frac{\log (2n)}{2\left(1-\rho-p_1-p_i-\sqrt{\varepsilon}\right)^2}\right\} +  \frac{C_{19}}{\varepsilon^2} + \frac{C_{20}}{\varepsilon} \right)\left(\Delta_i+2\Gamma_{i, \max}^2\right). \nonumber\end{equation}
\end{theorem}
Let $\varepsilon = (\log n)^{-\frac{1}{4}}$ and $n\rightarrow +\infty$, the regret bound in {Theorem~\ref{thm: thm4}} follows from Theorem \ref{thm: S-4}.

\end{document}